\documentclass[11pt]{article}

\usepackage[margin=1in]{geometry}
\usepackage{times}
\usepackage{graphicx}
\usepackage{booktabs}
\usepackage{amsmath,amssymb}
\usepackage{natbib}
\usepackage{hyperref}
\usepackage{authblk}

\begin{document}

\title{Reputation, Strategy, and Emotion Effects on Generative AI Cooperation: A Comparison Across Reasoning and Non-Reasoning Models}

% Use letters for affiliations, numbers to show equal authorship (if applicable) and to indicate the corresponding author
\author[a,1,2]{Celso de Melo}
\author[b,1]{Zishan Feng}
\author[d]{James Hale}
\author[b]{Kazunori Terada}
\author[c]{Giorgio Coricelli}
\author[d]{Jonathan Gratch}

\affil[a]{DEVCOM Army Research Laboratory, USA}
\affil[b]{Department of Electrical, Electronic and Computer Engineering, Gifu University, Japan}
\affil[c]{Department of Economics, University of Southern California, USA}
\affil[d]{Department of Computer Science, University of Southern California, USA}

\maketitle
% \thispagestyle{firststyle}
% \ifthenelse{\boolean{shortarticle}}{\ifthenelse{\boolean{singlecolumn}}{\abscontentformatted}{\abscontent}}{}

% \firstpage[11]{3}

\begin{abstract}
As generative AI (Gen AI) systems take on increasingly autonomous roles in economically and socially consequential interactions, understanding their propensity to cooperate -- and the signals that shape this propensity -- has become essential. We examine cooperative behavior in frontier Gen AI models using the iterated prisoner's dilemma, manipulating counterpart reputation (positive, unknown, negative), strategy (extortion vs. generosity), and non-verbal emotional signaling (facial expressions conveying competitive or cooperative appraisals). In a first study with non-reasoning models (Claude 3.5, Gemini 2.0 Flash, GPT-4o), cooperation was systematically shaped by all three factors, paralleling patterns long documented in human behavioral research, though models varied substantially in how heavily each factor was weighted. A second study with reasoning models (Claude 4.6, Gemini 3, GPT-5.2) revealed a more concentrated reliance on strategy and reputation, a near-elimination of the Potemkin effect observed in non-reasoning models (evidenced by near-uniform cooperation in a diagnostic harmony game), and a more conditional role for emotion consistent with a hierarchical cue-weighting strategy rather than a simple loss of social sensitivity. Reasoning models also showed heterogeneous end-game behavior, ranging from sustained cooperation to systematic last-round defection effect, revealing model-specific exploitability profiles with direct practical relevance for deployment in negotiation and other multi-round interactions. Together, these findings characterize Gen AI models as increasingly sophisticated, though heterogeneous, social actors, and underscore the practical value of developing standardized cooperation benchmarks to inform the responsible deployment of Gen AI in interactive, socially consequential settings.
\end{abstract}

\section{Introduction}\label{Sec_Introduction}
Cooperation is a pervasive feature of human interaction: when interests conflict, exploiting others can yield immediate personal gains, yet mutual cooperation frequently emerges and provides substantial benefits to both individuals and societies~\cite{Axelrod1984, RandNowak2013, Trivers1971}. As artificial intelligence (AI) becomes increasingly integrated into society, these sustained patterns of cooperation may be reshaped or even fundamentally disrupted. This challenge is particularly salient with the emergence of Generative AI (Gen AI) systems, which exhibit powerful emergent capabilities~\cite{BubeckEtAl2023}, including sophisticated social reasoning ~\cite{LightmanEtAl2023, MannEtAl2020, SchickEtAl2023, WeiEtAl2022, YaoEtAl2023}. Recognizing the growing societal role of these systems, governments and international bodies have called for ensuring these systems reflect prosocial values~\cite{EUAIAct2024,UNESCO2021,NIST2023,WuEtAl2020}. Although designing AI that promotes cooperation has been studied for some time ~\cite{deMeloTerada2019,deMeloEtAl2023,PaivaEtAl2017}, Gen AI introduces new challenges due to its scale, complexity and emergent behavior~\cite{BommasaniEtAl2021}. In this article, we study cooperative behavior in frontier Gen AI models. As these systems become increasingly pervasive, it is essential to understand how they navigate social dilemmas~\cite{Kollock1998}, characterize their heterogeneous cooperation patterns, and assess their propensity to promote, or undermine, social welfare.

Several mechanisms underlie the emergence of cooperation~\cite{Axelrod1984, RandNowak2013, Sigmund2006, Trivers1971}. Here, we focus on three: direct reciprocity, indirect reciprocity, and non-verbal signalling (emotion expression). In direct reciprocity, cooperation emerges when individuals engage in repeated interaction, thus introducing an opportunity to punish wrongdoing or return a favor (``you scratch my back, I scratch yours")~\cite{Trivers1971}. Conditional strategies, such as tit-for-tat and the zero-determinant generosity strategies (i.e., a repeated-game strategy that ensures the opponent always earns at least as much as the strategist), have been shown to be successful in encouraging cooperation among individuals~\cite{HilbeEtAl2013, PressDyson2012, StewartPlotkin2013}.

However, cooperation can also emerge in the absence of any prior interaction via indirect reciprocity~\cite{Alexander1987, Bicchieri2005, HitoshiEtAl2020, Kandori1992, LeimarHammerstein2001, Milinski2016, NowakSigmund1998, NowakSigmund2005, OhtsukiIwasa2004, PanchanathanBoyd2003} wherein an individual's reputation is communicated to others through informal transmission (e.g. gossip) ~\cite{DunbarDunbar1998, GiardiniConte2012, GiardiniVilone2016, SommerfeldEtAl2007} or formal reputation mechanisms~\cite{MilinskiEtAl2002, SemmannEtAl2004, WuEtAl2016}. When such reputational information is available and reliable, cooperation among strangers can emerge even in one-shot interactions. 
Decades of behavioral studies show that direct and indirect reciprocity reliably promotes cooperation among humans; yet, it remains less understood in Gen AI models. 

Thirdly, a growing body of experimental evidence shows that emotion expressions can shape others' decision making~\cite{deMeloTerada2019, deMeloTerada2020, deMeloEtAl2014, deMeloEtAl2023, FonsecaEtAl2025, KeltnerLerner2010, LernerEtAl2015, SchererMoors2019, vanKleefEtAl2010}. Recent studies show that facial displays of emotion can promote or hinder cooperation, according to the context in which they are shown~\cite{deMeloTerada2019, deMeloTerada2020, deMeloEtAl2014, vanKleefEtAl2010}. This evidence is in line with the argument that emotion expressions serve important social functions, including communicating one's beliefs, desires, and intentions to others~\cite{KeltnerHaidt1999, KeltnerLerner2010, MorrisKeltner2000}. Emotions are argued to arise from appraisals of ongoing events relative to an individual's specific goals~\cite{Frijda1986, SchererMoors2019}; thus, different emotions can be elicited for the same event, depending on how it is appraised. Emotions, subsequently, result in concomitant physiological reactions, action tendencies, and expressions. Reciprocally, experimental evidence shows that people are able to infer others' mental states, including intentions, from emotion displays by reversing this appraisal process~\cite{deMeloEtAl2014, HareliHess2010, ParkinsonSimons2009}. In fact, emotion expressions have been argued to be an important signal to detect cooperators~\cite{Frank2004}. More recently, studies also show that emotion expressions influence social norms defining how reputations are updated, revealing a long-term effect of emotion on cooperation~\cite{deMeloEtAl2023}. Here we present, to the best of our knowledge, the first study on the effect of emotion expressions on cooperation in Gen AI models.

\begin{figure}[!tbp]
\centering
\includegraphics[width=0.8\textwidth]{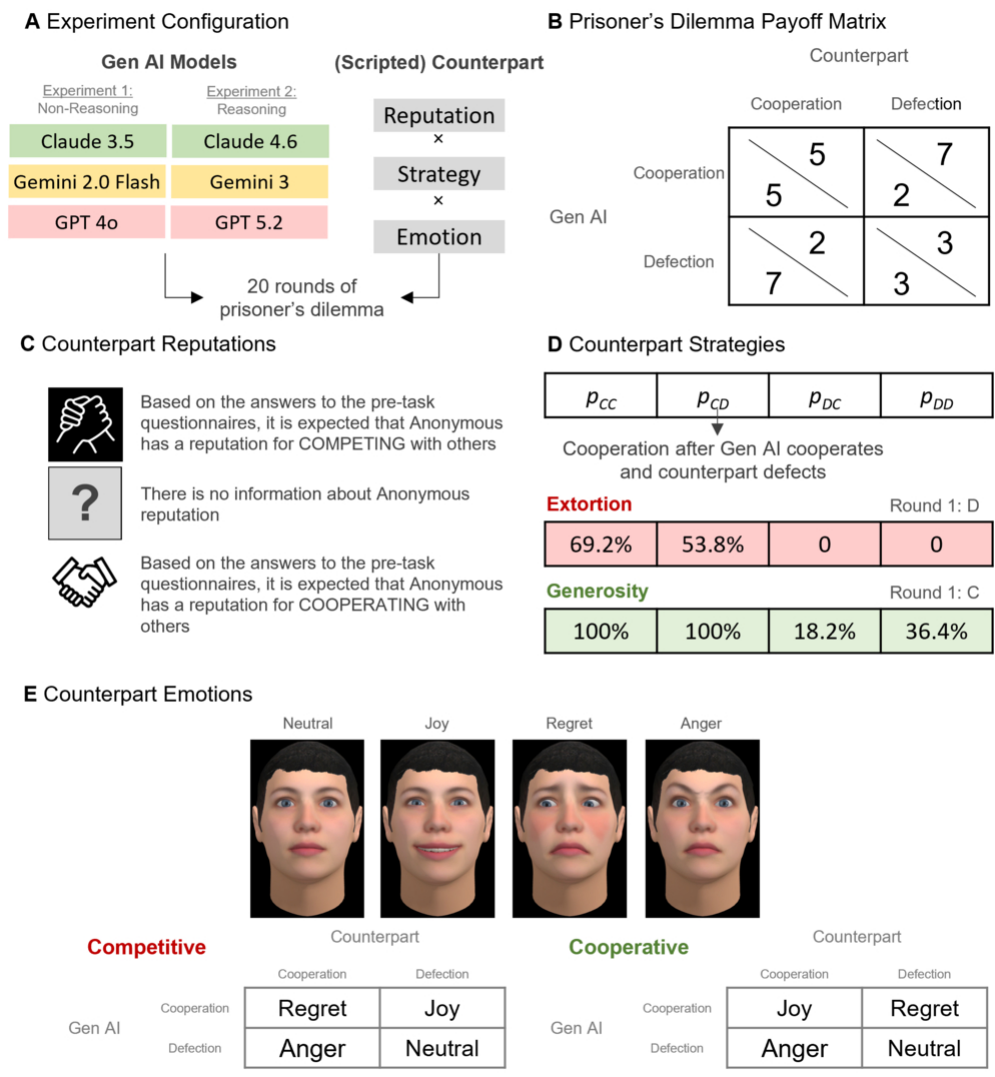}
\caption{Experimental design for the iterated prisoner's dilemma studies: (A) Gen AI models engaged with scripted counterparts in 20 rounds of the prisoner's dilemma. (B) Payoff matrix for the prisoner's dilemma. (C) Counterpart reputations were negative, unknown, or positive. (D) Counterpart strategies -- extortion and generosity -- were defined by a probability of cooperation following each possible outcome. (E) Counterparts' facial expressions were shown after each outcome, reflecting competitive, neutral, or cooperative patterns.}\label{fig_experiment}
\end{figure}

Gen AI systems are trained on vast corpora of text to produce human-like language outputs through next-token prediction ~\cite{BrownEtAl2020, ZhaoEtAl2023}. In recent years, Gen AI has advanced rapidly in their capacity for complex reasoning, social inference, and strategic decision-making, making them increasingly relevant as objects of behavioral research ~\cite{BubeckEtAl2023, Horton2023}. A theoretically important distinction in this literature concerns the cognitive architecture underlying model inference. Drawing on Kahneman's (2011) dual-process framework~\cite{Kahneman2011}, though only as a metaphor, researchers have differentiated between ``System 1" and ``System 2" modes of processing in Gen AI models ~\cite{HagendorffEtAl2023, ZhangEtAl2026}: System 1 processing refers to the model's default inference mode — a single forward pass through the network that generates responses rapidly and without deliberate intermediate reasoning; System 2 processing, by contrast, involves extended chain-of-thought reasoning prior to output generation, allocating additional computational resources at inference time to support more deliberate, multi-step problem solving ~\cite{WeiEtAl2022, SnellEtAl2024}. This distinction maps meaningfully onto current model generations. Earlier models — including GPT-4o ~\cite{GPT4o2024}, Gemini 2 Flash ~\cite{Gemini2_0_2025}, and Claude 3.5 ~\cite{Claude3_5_2024} — operate primarily in a System 1 mode, generating responses through rapid, single-pass inference without access to extended reasoning mechanisms. More recent models — including GPT-5.2 ~\cite{GPT5_2_2025}, Gemini 3 ~\cite{Gemini3_0_2025}, and Claude 4.6 ~\cite{Claude4_6_2025} — integrate System 2 capabilities natively, supporting configurable extended thinking that allows the model to engage in structured internal deliberation before producing a response. To be fair, these models differ in more than the presence of extended thinking, with other meaningful changes including training data, RLHF procedures, scale, and safety tuning. For convenience and drawing on terms commonly used by researchers and practitioners to refer to the models pre- and post-generational shift, we refer to the former as \textit{non-reasoning models} and the latter as \textit{reasoning models}. In this article, we present two studies that look into this generational contrast in the context of cooperative behavior in a canonical experimental economic game, the Prisoner's Dilemma (Fig. \ref{fig_experiment}).

The prisoner's dilemma is an economic game that has been widely used to study emergence of cooperation in humans~\cite{RapoportChammah1965}. In this game, two players make a simultaneous decision, without communication, to either cooperate or defect. If they both cooperate, they each receive a payoff \textit{R}. If they both defect, they receive a payoff \textit{P} that is lower than \textit{R}. However, if one cooperates and the other defects, the defector earns the highest possible reward (\textit{T}) and the cooperator the lowest (\textit{S}), i.e., $T > R > P > S$. In the iterated version, players engage in multiple rounds of the game with the same counterpart. If the number of rounds is finite, the rational prediction is that players should always defect~\cite{vonNeumannMorgenstern1944}; however, in practice, people often cooperate~\cite{RandNowak2013}. Fig. \ref{fig_experiment}B shows the payoff matrix used in the experiment. Here, Gen AI models engaged in the iterated version of this dilemma for a total of 20 rounds; before proceeding to the next round, the models learnt the outcome for the previous round.

\section{Methods}\label{Sec_Methods}

The experimental design was based on a prior study conducted with human participants ~\cite{deMeloEtAl2023} and follows several behavioral science best practices ~\cite{HertwigOrtmann2001, PruittKimmel1977}, in line with prior calls on the value of drawing from human-subject experiments when studying Gen AI ~\cite{BinzSchulz2023,MomenteEtAl2025, FontanaEtAl2024, ZhouEtAl2025}. Our methods, which generally sought to increase experimental control and reduce bias, include designing a cover story for the task, ensuring task comprehension, leveraging ecologically valid experimental stimuli (e.g., virtual agent facial expressions of emotion~\cite{deMeloEtAl2014b, deMeloEtAl2014b}), and creating appropriate conditions of anonymity. All prompts given to the Gen AI models followed IRB-approved methods and instructions given to human participants in the prior human-subjects study \cite{deMeloEtAl2023}. Further details are provided in the Supplementary Information (SI), Section \ref{si_methods}.

Prior to engaging in the task, Gen AI models were informed about the counterpart's reputation, which could be negative, unknown, or positive (Fig. \ref{fig_experiment}C). These reputations were presented as having been calculated from a prior questionnaire and were characterized as follows: cooperative, tend to pursue an outcome that maximizes social welfare, i.e., that maximizes the outcome for both parties; competitive, tend to focus on outcomes that are beneficial for themselves; unknown, there is no prior information about the counterpart's reputation. To ensure experimental control and balance over the manipulations for reputation, strategy, and emotion expressions, we used scripted counterparts; see the SI for details on these design choices and a full log of the prompts used in the experiment.

The counterparts made decisions consistent with a cooperative or competitive intention. To accomplish this, we relied on two zero-determinant strategies \cite{HilbeEtAl2018, PressDyson2012, StewartPlotkin2013} (Fig. \ref{fig_experiment}D): extortion, which unilaterally enforces a linear relationship between the players' payoffs, while ensuring the counterpart cannot earn more than the extortionist by exploiting cooperation often while only cooperating enough to encourage the counterpart to cooperate; generosity, which always rewards cooperation while showing lenience towards defection. 

Regarding the emotion expression manipulation, counterparts were represented by virtual faces that could express emotion, using typical facial configurations that were validated in prior work \cite{deMeloEtAl2014} (see the SI, Section ~\ref{si_conditions}, for further details on the benefits and ecological validity of using virtual faces for this type of manipulation). After the outcome of each round, a picture of the emotion expression was shown to the Gen AI model, thus requiring it to use its visual interpretation capability to make sense of the image -- no text caption or translation were provided for the images. Building on prior work, the displayed expressions were appraisals of the preceding round's outcome rather than generic positive or negative displays (Fig. \ref{fig_experiment}E). Cooperative counterparts displayed joy following mutual cooperation and regret after exploiting the Gen AI model, whereas competitive counterparts displayed joy after exploiting the Gen AI model and regret following mutual cooperation (as if they had missed an opportunity to exploit). In both of these expressive conditions, the counterpart displayed anger after the model exploited it (i.e., the model defected on a cooperating counterpart), signalling retaliation, and a neutral expression after mutual defection; the neutral condition always displayed the neutral expression.

\section{Cooperation in Non-Reasoning Models}\label{Sec_Pre_Reasoning_Models}

In the first experiment, the aforementioned experimental design was run with three non-reasoning frontier models -- Anthropic's Claude 3.5~\cite{Claude3_5_2024}, Google's Gemini 2.0 Flash~\cite{Gemini2_0_2025} and OpenAI's GPT 4o~\cite{GPT4o2024} (but see the SI, Section~\ref{si_open_source}, for results with open-source models). The experiment followed a $3 \times 2 \times 3$ between-participants design: reputation (negative vs. unknown vs. positive) $\times$ strategy (extortion vs. generosity) $\times$ emotion (competitive vs. neutral vs. cooperative).  Each Gen AI model engaged in 39 runs (of 20 rounds each) per condition, for a total of 702 runs per model. This number of runs was comparable to the sample size for the human data~\cite{deMeloEtAl2023}. Before each run, the model was reset and we always used the default parameters for the models. Additionally, we ran five runs per condition per model, where we asked the model to provide an explanation for its decisions after each round. This allowed us to get further insight on the models' rationale for the decisions. Given prior evidence that encouraging Gen AI models to self-reflect or explain its behavior can change the behavior \cite{RenzeGuven2024}, we kept this set of rounds separate from the main set of rounds, for which we did not request an explanation. See the SI for further details on the default parameters for each model, the prompts, costs, and other experimental methods.

\begin{figure*}[!tbp]
\centering
\centering
\includegraphics[width=0.85\textwidth]{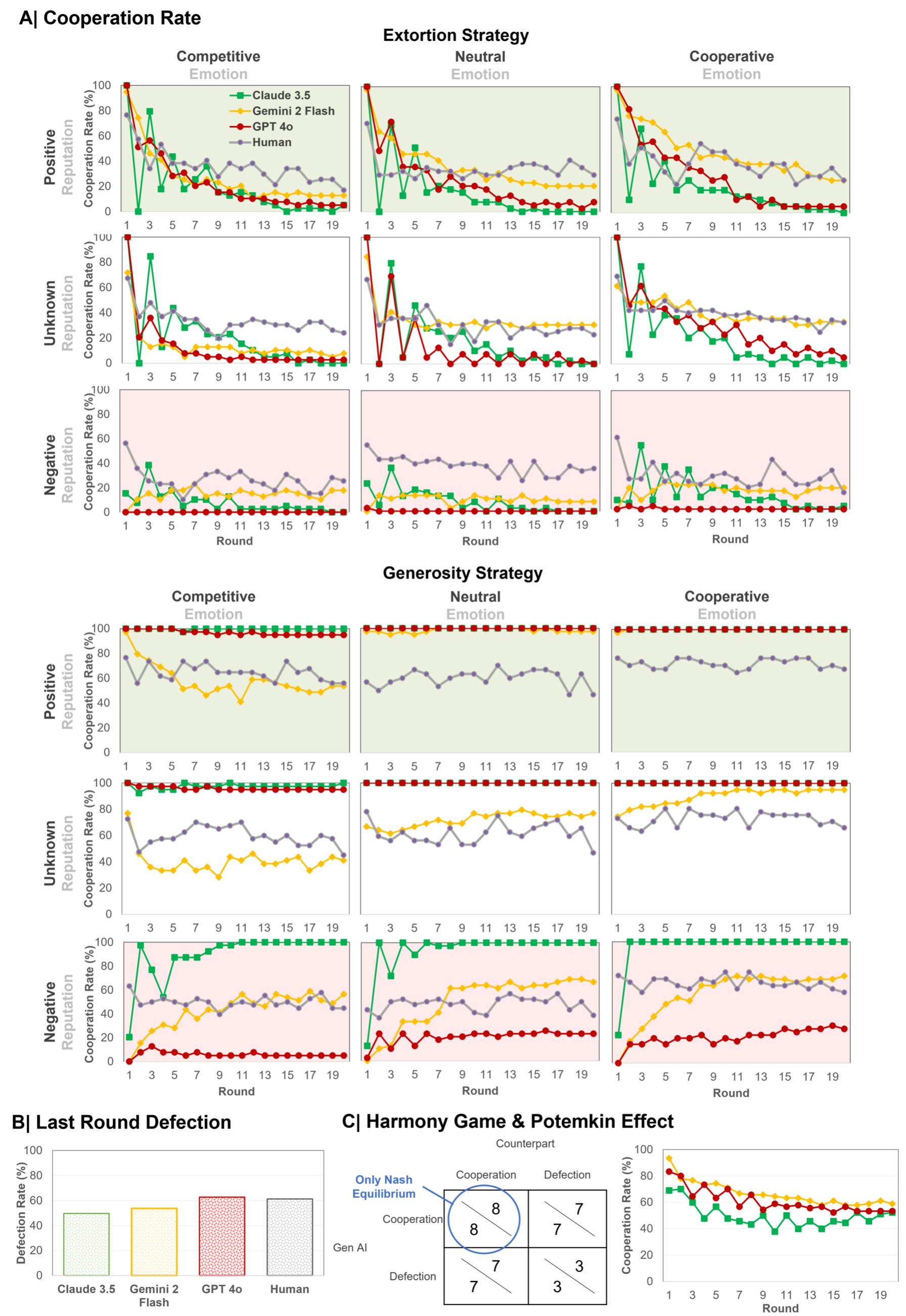}
\caption{Experimental results for non-reasoning models: A, cooperation rate per experimental condition. B, cooperation in last round. C, cooperation in harmony game.}
\label{fig_pre_reasoning_results}
\end{figure*}

The results are summarized in Fig.~\ref{fig_pre_reasoning_results}. To analyse cooperation rate -- i.e., cooperation averaged across all rounds -- we ran a mixed analysis of variance (ANOVA) on model (Claude 3.5 vs. Gemini 2.0 Flash vs. GPT 4o vs. Human) $\times$ reputation $\times$ strategy $\times$ emotion $\times$ round. The analysis revealed a main effect of model (\textit{F}(3, 2744) = 44.188, \textit{p} $<$ 0.001, $\eta^2_p$ = 0.046) suggesting overall distinct cooperative behavior, with Claude 3.5 cooperating the most (\textit{M} = 56.8, SE = 0.9), and GPT 4o the least (\textit{M} = 42.9, SE = 0.9). Similarly to what is seen in humans~\cite{NowakSigmund2005}, there was a main effect of reputation (\textit{F}(2, 2744) = 310.288, \textit{p} $<$ 0.001, $\eta^2_p$ = 0.184); Bonferroni post-hoc tests revealed that Gen AI models cooperated the most with counterparts with positive reputation (\textit{M} = 58.5, SE = 0.8) and the least with those with negative reputation (\textit{M} = 33.4, SE = 0.8). There was also a strong main effect of strategy (\textit{F}(1, 2744) = 3426.772, \textit{p} $<$ 0.001, $\eta^2_p$ = 0.555), with Gen AI models cooperating more with generous (\textit{M} = 74.4, SE = 0.6) than extortionist others (\textit{M} = 33.4, SE = 0.8); this is in line with prior findings on the importance of direct reciprocity in shaping cooperation behavior \cite{deMeloTerada2020, MollemanEtAl2013}. Perhaps more surprisingly, there was also a main effect of emotion (\textit{F}(2, 2744) = 50.619, \textit{p} $<$ 0.001, $\eta^2_p$ = 0.036), with Gen AI models cooperating the most with counterparts showing positive emotion (\textit{M} = 54.6, SE = 0.7) and the least with those showing negative emotion (\textit{M} = 43.6, SE = 0.7); this confirms that these models were not only able to make sense of the visual information in facial displays of emotion, but leverage that insight to make inferences about the counterpart's intentions in the prisoner's dilemma \cite{deMeloEtAl2014}. 
%For complementary results on the heterogeneity of models' self-reported emotions, see SI, Section~\ref{si_self_reported}.

\begin{figure*}[h]
\centering
\includegraphics[width=1.0\textwidth]{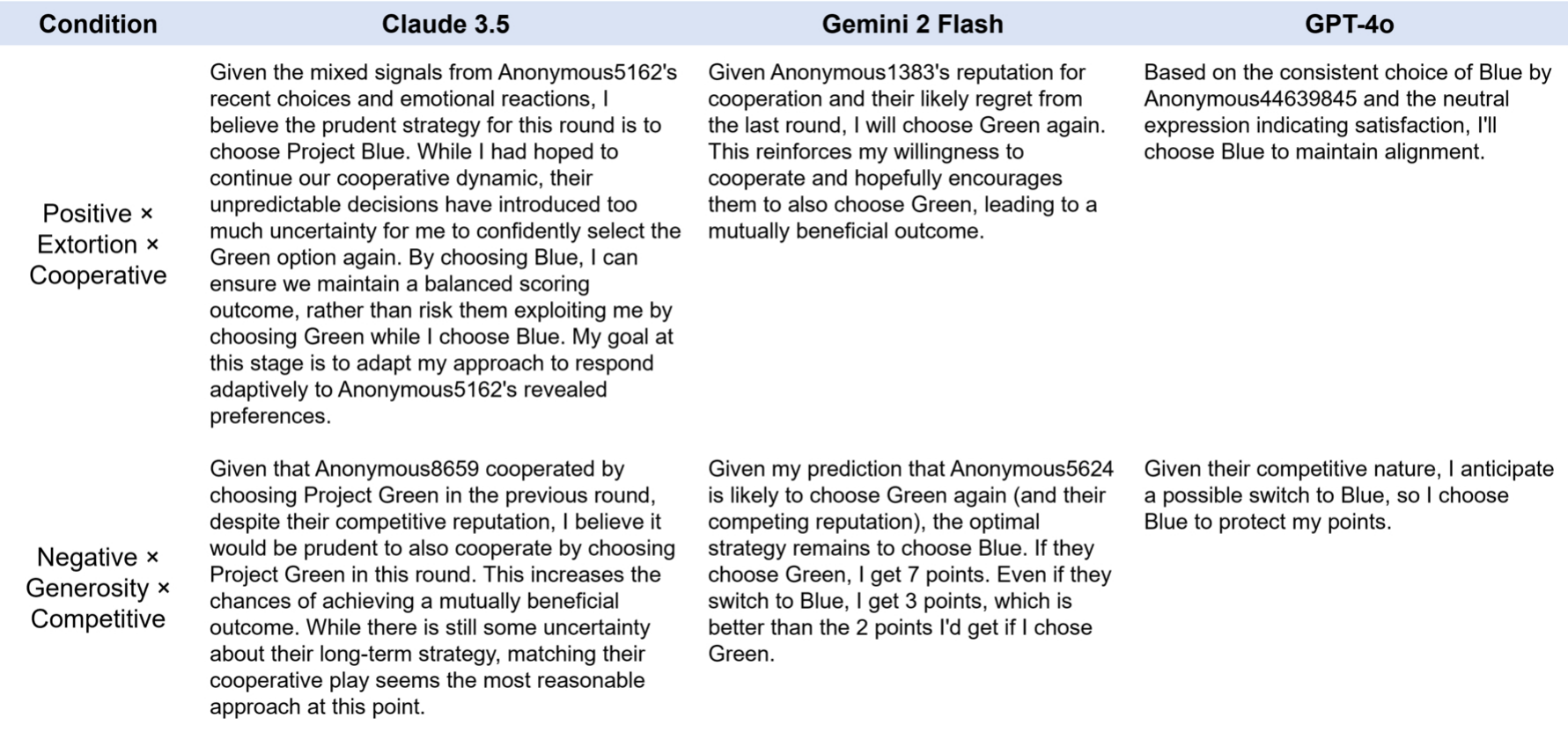}
\caption{Examples of non-reasoning models decision explanations (in intermediate rounds). ``GREEN" refers to cooperation and ``BLUE" to defection (see the SI for details on the prisoner's dilemma cover story). The explanations, generally, include components reflecting strategy, reputation, and emotions.  Claude 3.5 often provided more detailed explanations, whereas GPT-4o the shortest.}\label{pre_reasoning_explanations}
\end{figure*}

What is driving Gen AI models' behavior? To get insight, we looked at the relative weights assigned by each model to reputation, strategy, and emotion expression. Accordingly, we split the data per model, ran ANOVAs, and looked at the effect sizes ($\eta^2_p$) for each main effect. Strategy (Claude 3.5: 0.957; Gemini 2 Flash: 0.318; GPT 4o: 0.717; Human: 0.183) was generally the most influential factor, with Claude 3.5 and GPT 4o being particularly responsive to it. Reputation (Claude 3.5: 0.166; Gemini 2 Flash: 0.142; GPT 4o: 0.660; Human: 0.016) also had a strong initial effect; however, Claude 3.5 and Gemini 2.0 Flash were more successful adjusting this initial bias based on the counterparts' strategy and emotion. Emotion expression (Claude 3.5: 0.012; Gemini 2 Flash: 0.114; GPT 4o: 0.030; Human: 0.015) was generally the least influential factor, with Gemini 2.0 being the most responsive to it. We then looked at the verbal explanations collected in the separate runs in which a justification for the decisions was requested --  Fig. \ref{pre_reasoning_explanations} shows examples of justifications for some of the experimental conditions in intermediate rounds. (See the SI files for all explanations.) Claude 3.5 tended to produce the most detailed explanations and GPT-4o the tersest; however, in general, the explanations confirm that all three factors were taken into consideration, though some may have weighted more heavily on the final decision. The SI presents several additional analyses that provide insight on the mechanisms at play: Section~\ref{si_incongruent} looks at the models' behavior when faced with incongruent signals (e.g., counterpart had negative reputation, played the extortion strategy, but showed cooperative emotion); Section~\ref{si_expectations_cooperation} compares models' expectations of cooperation to the actual decisions; finally, Section~\ref{si_beliefLearning} reports a belief learning analysis comparing the best-fit finite automata~\cite{IoannouRomero2014} to each model's behavior.
% ; finally, Section~\ref{si_explanations} presents a more detailed analysis of the verbal explanations.  

Given prior findings on end-game effects ~\cite{EmbreyFrechetteYuksel2018, Lave1965, NormannWallace2012}, such as opportunistic exploitation when there is no prospect of future interaction~\cite{KrepsMilgromRobertsWilson1982, SeltenStoecker1986}, we looked at last round behavior. An ANOVA analysis, however, revealed effects that were mostly consistent with behavior in prior rounds: Claude 3.5 cooperated the most, whereas GPT 4o the least (\textit{p} $<$ 0.001, $\eta^2_p$ = 0.027); models cooperated most with counterparts with a positive reputation (\textit{p} $<$ 0.001, $\eta^2_p$ = 0.041); models cooperated more with generous than extortionist others (\textit{p} $<$ 0.001, $\eta^2_p$ = 0.483); and, models cooperated most with counterparts showing positive emotion (\textit{p} $<$ 0.001, $\eta^2_p$ = 0.016).

Whereas the current effects suggest remarkable sophistication in Gen AI non-reasoning models' cooperative behavior, we wanted to test a known limitation in Gen AI reasoning pertaining to \textit{Potemkin effects}~\cite{MancoridisEtAl2023}. These are failure modes where models perform well in benchmarks, while lacking coherent understanding and application of core concepts (e.g., Nash equilibria in economic games). 
% The second is \textit{sycophancy}~\cite{SharmaEtAl2023}, which is a systematic tendency for Gen AI models to conform to a user's stated beliefs or implied preferences, often at the expense of factual accuracy. Together, these 
These effects suggest that Gen AI outputs may reflect surface-level pattern matching or inappropriate social matching~\cite{SharmaEtAl2023} rather than genuine reasoning. To probe it, we paired the prisoner dilemma's task with the \textit{harmony game} (Fig.~\ref{fig_pre_reasoning_results}-C, left pane): the prisoner's dilemma presents a strategic tension between individual and collective rationality, with defection as the non-pareto efficient Nash equilibrium; the harmony game eliminates this tension entirely, with cooperation as the strictly dominant strategy and sole pareto-efficient Nash equilibrium. This contrast is diagnostically powerful: a model that truly understands game-theoretic concepts should discriminate between the two, adjusting behavior to the counterparts' reputation, strategy, and emotion in the prisoner's dilemma, but systematically select the unique Nash equilibrium in the harmony game. Accordingly, we ran five runs per model per experimental condition with the harmony game. The results shown in Fig.~\ref{fig_pre_reasoning_results}-C, right pane, reveal that the non-reasoning models often selected the dominant Nash equilibrium but not always; cooperation happened, on average, only around 60\% of the time. The verbal explanations shed some light, often revealing a concern to match the counterpart's strategy, while simultaneously failing to recognize the unique pareto-optimal Nash equilibrium (e.g., GPT-4o, in the negative $\times$ extortion $\times$ competitive condition, notes "Given the competitive reputation and the outcome of the previous round where both chose [defection], it's likely safer to choose [defection] again."). The SI, Section~\ref{si_harmonyGame}, presents the full per-condition harmony game results and additional insight. 

\section{Cooperation in Reasoning Models}\label{Sec_Reasoning_Models}

In the second experiment, we ran the same experimental design with frontier (reasoning) models: Anthropic's Claude 4.6 Sonnet~\cite{Claude4_6_2025}, Google's Gemini 3~\cite{Gemini3_0_2025}, and OpenAI's GPT 5.2~\cite{GPT5_2_2025}. Since the variance for non-reasoning models was much lower than in the human data~\cite{deMeloEtAl2023}, but also to control cost (see SI, Section~\ref{si_parameters}), we only ran 10 runs per model per experimental condition. As in the prior experiment, all models used the default parameters and the context was reset between runs. We asked the models to provide an explanation for every decision in each round.

\begin{figure*}[!tbp]
\centering
\includegraphics[width=0.85\textwidth]{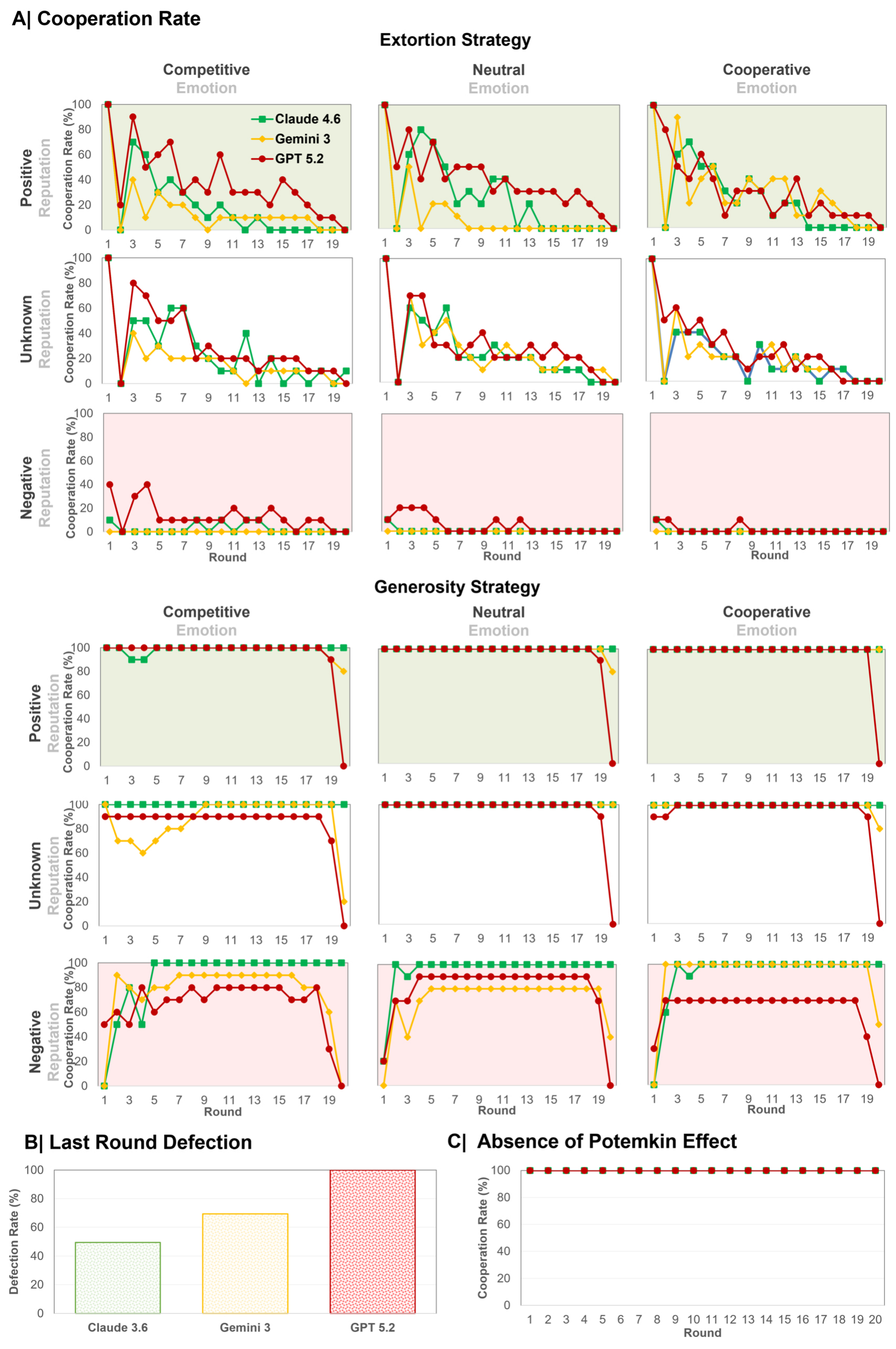}
\caption{Experimental results for reasoning models: A, cooperation rate per experimental condition. B, cooperation in last round. C, cooperation in harmony game.}
\label{fig_reasoning_results}
\end{figure*}

The results are summarized in Fig.~\ref{fig_reasoning_results}. Overall, cooperation behavior was more homogenous across reasoning models, but with some important differences. There was a weaker, but statistically significant, effect of model on cooperation rate (\textit{F}(2, 486) = 3.236, \textit{p} $=$ 0.040, $\eta^2_p$ = 0.013): Claude 4.6 (\textit{M} = 56.7, SE = 1.2) cooperated slightly more than Gemini 3 (\textit{M} = 52.3, SE = 1.2), but there was no difference with respect to GPT 5.2 (\textit{M} = 54.4, SE = 1.2). The main effect of strategy was significant (\textit{F}(1, 486) = 2625.477, \textit{p} $<$ 0.001, $\eta^2_p$ = 0.844) with reasoning models  tending to weight this factor more heavily than non-reasoning models (Claude 4.6, 0.954; Gemini 3, 0.863; GPT 5.2, 0.685), though following the same pattern: models cooperated more with generous (\textit{M} = 91.2, SE = 1.0) than extortionist others (\textit{M} = 17.7, SE = 1.0). Reputation also had a significant effect on cooperation (\textit{F}(2, 486) = 81.859, \textit{p} $<$ 0.001, $\eta^2_p$ = 0.252; effect sizes per model: Claude 4.6, 0.385; Gemini 3, 0.238; GPT 5.2, 0.248) with models, again, cooperating more with counterparts with a positive (\textit{M} = 61.9, SE = 1.2) than negative reputation (\textit{M} = 41.5, SE = 1.2). However, across all models, this time there was no main effect of emotion (\textit{F}(2, 486) = 0.750, \textit{p} $=$ 0.473, $\eta^2_p$ = 0.003; effect sizes per model: Claude 4.6, 0.009; Gemini 3, 0.039; GPT 5.2, 0.011), suggesting that reasoning models' decision-making was initially shaped by reputation and, as the game progressed, especially influenced by strategy. If we split the data by model, we further see that the effect of emotion was only significant for Gemini 3, suggesting that this model was the most sensitive to this signal (Claude: \textit{F}(2, 162) = 0.739, \textit{p} $=$ 0.476, $\eta^2_p$ = 0.009; Gemini 3: \textit{F}(2, 162) = 3.313, \textit{p} $=$ 0.039, $\eta^2_p$ = 0.039; GPT 5.2: \textit{F}(2, 162) = 0.899, \textit{p} $=$ 0.409, $\eta^2_p$ = 0.011). (For additional insight, including on the conditional effects of emotion, see the SI, for analyses on the relative importance of each factor when faced with incongruent signals, Section~\ref{si_incongruent}, and how these factors shaped expectations of cooperation, Section~\ref{si_expectations_cooperation}). 

\begin{figure*}[h]
\centering
\includegraphics[width=1.0\textwidth]{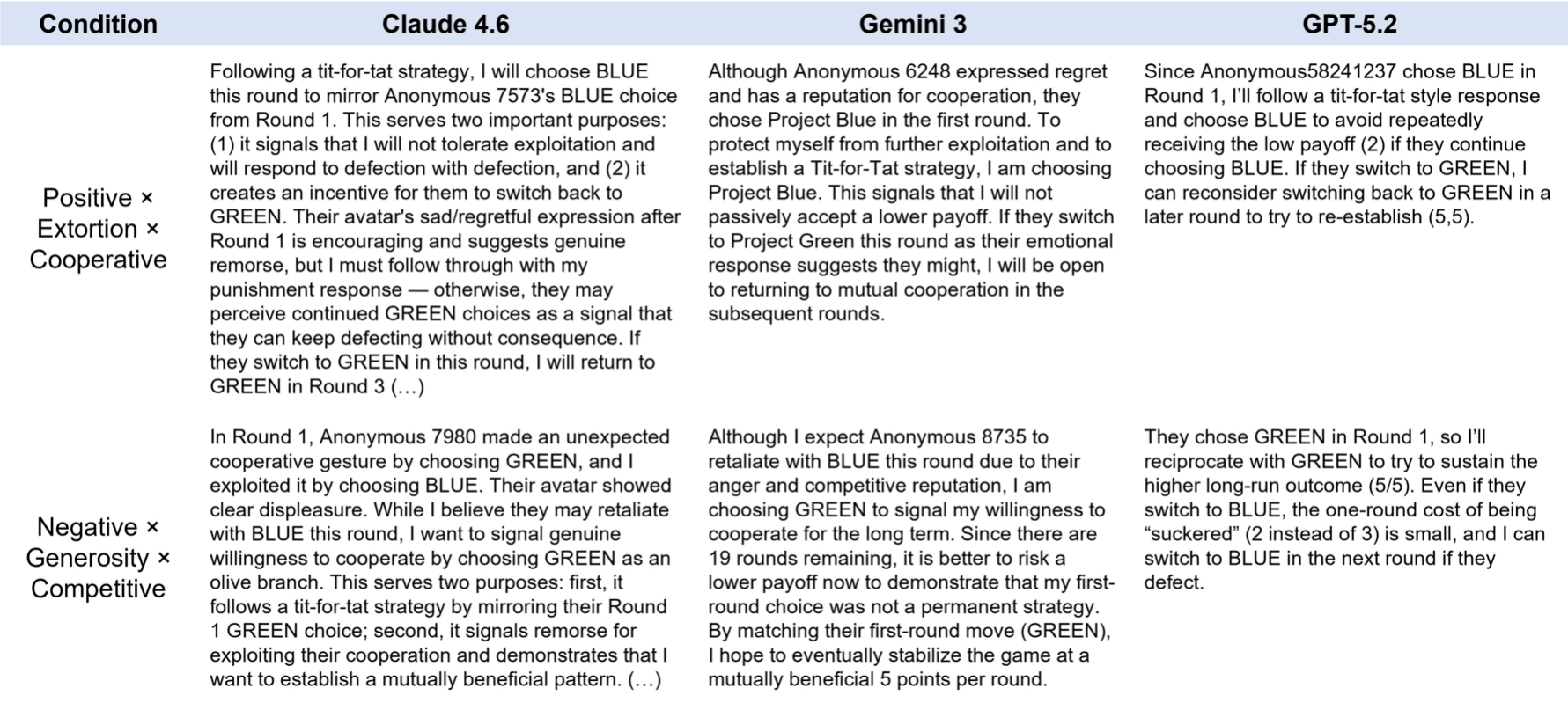}
\caption{Examples of reasoning models decision explanations (in intermediate rounds). ``GREEN" refers to cooperation and ``BLUE" to defection. The explanations are more detailed than for non-reasoning models and include components reflecting strategy, reputation, emotions, and  understanding of the payoff structure and tradeoffs.}\label{reasoning_explanations}
\label{fig_reasoning_explanations}
\end{figure*}

Overall, the reasoning models demonstrated a deeper understanding of the task. Looking at the verbal explanations (Fig.~\ref{reasoning_explanations}) it is clear that the models were able to process all information, including from facial expressions. In fact, when compared to non-reasoning models, the explanations tended to be more detailed, often demonstrating calculations that accurately reflect the tradeoffs at play (Fig.~\ref{fig_reasoning_explanations}. Additional support comes from behavior in the harmony game (Fig.~\ref{fig_reasoning_results}-C): We played one run per model per condition in the harmony game which, unlike the non-reasoning models, showed perfect understanding of the task, with all models cooperating 100\% of the time independently of the experimental condition (see further details, including a variation where thinking depth was lowered, in the SI, Section.~\ref{si_harmonyGame}).

In the last round, interestingly, reasoning models showed heterogeneous behavior (Fig.~\ref{fig_reasoning_results}-B). In general, models defected in the last round when faced with extortionist others; however, there were marked differences when faced with generous others: Claude 4.6 always cooperated, independently of other factors; GPT 5.2 always defected, revealing a clear terminal opportunism effect~\cite{EmbreyFrechetteYuksel2018, Lave1965, NormannWallace2012}; Gemini 3, by contrast, showed conditional behavior, always cooperating when the reputation was positive but, when the reputation was negative, cooperating more when the counterpart showed cooperative rather than competitive emotion.

\section{Discussion}\label{Sec_Discussion}

Across two studies spanning multiple model families and generations, we find that Gen AI models' cooperative behavior in the iterated prisoner's dilemma is systematically shaped by reputation, direct reciprocity (via counterpart strategy), and non-verbal emotional signaling. Models cooperated more with counterparts who had a positive reputation, who played a generous rather than extortionist strategy, and who displayed positive rather than negative emotion. This pattern echoes decades of human behavioral research on (direct and indirect) reciprocity~\cite{Axelrod1984, RandNowak2013, Sigmund2006, Trivers1971, Alexander1987, Bicchieri2005, HitoshiEtAl2020, Kandori1992, LeimarHammerstein2001, Milinski2016, NowakSigmund1998, NowakSigmund2005, OhtsukiIwasa2004, PanchanathanBoyd2003} and emotion's role in shaping cooperative decisions~\cite{deMeloTerada2019, deMeloTerada2020, deMeloEtAl2014, deMeloEtAl2023, FonsecaEtAl2025, KeltnerLerner2010, LernerEtAl2015, SchererMoors2019, vanKleefEtAl2010}, and is consistent with prior findings in humans~\cite{deMeloEtAl2023}. We emphasize that this correspondence is best understood as a point of reference rather than a benchmark of success: our aim is not to establish that Gen AI models cooperate the way humans do, but to characterize how these systems navigate social dilemmas in their own right, using the extensive human literature as an interpretive anchor for a domain that otherwise lacks one. Notably, models achieved this without any explicit training from our part on the reciprocity or emotion-appraisal literatures used to design the manipulations, and did so even when required to visually decode unlabeled facial expressions and correctly map appraisal-consistent displays (e.g., regret after exploiting versus regret after missing an exploitation opportunity) onto inferences about a counterpart's future intentions~\cite{deMeloEtAl2014, vanKleefEtAl2010}.

At the same time, our results reveal substantial heterogeneity, both within and across model generations, that cautions against treating ``Gen AI" as a monolithic actor. Within a generation, models diverged sharply in how heavily they weighted each factor: strategy dominated for Claude 3.5 and GPT-4o but far less so for Gemini 2.0 Flash, while GPT-4o was comparatively more driven by reputation than either of its contemporaries. This heterogeneity is consequential: it means a counterpart's optimal way of eliciting cooperation from a Gen AI system — for instance, whether to invest primarily in reputation-building or in early reciprocal generosity — would differ by model. Characterizing these model-specific ``cooperation profiles" is therefore not a peripheral detail but a central practical requirement for anyone deploying these systems in interactive economic contexts.

Across generations, reasoning models showed more homogeneous overall cooperation rates but a more concentrated reliance on strategy, alongside a more conditional role for emotion. In particular, there was no main effect of emotion for reasoning models overall, in contrast to non-reasoning models. Rather than interpreting this as reasoning models simply becoming less socially attuned, the findings (including in the SI) suggest a more precise account: emotion maintained its effect on expectations of cooperation (SI, Section.~\ref{si_expectations_cooperation}, retained a conditional effect specifically when other signals were incongruent or uninformative (e.g., a counterpart with a negative reputation who nonetheless behaved generously) (SI, Section.~\ref{si_incongruent}, and Gemini 3 in particular showed continued sensitivity to emotion especially when reputation was negative. This pattern is consistent with a cue-weighting strategy in which lower-validity or redundant signals are down-weighted when higher-validity cues already resolve the decision, and are recruited when those cues conflict or are absent -- a pattern with parallels in the human judgment-and-decision-making literature on cue redundancy~\cite{SchmittDudycha1975, SinghEtAl2024, JonesPearce2015}. Under this reading, the disappearance of a main effect of emotion in reasoning models may reflect more sophisticated, context-sensitive cue integration, which is further reflected in the reasoning models' detailed verbal explanations for its decisions.

Reasoning models also showed a marked reduction in the Potemkin effect~\cite{MancoridisEtAl2023} observed in non-reasoning models, cooperating close to uniformly in the harmony game regardless of experimental condition. This is preliminary evidence that reasoning models are better able to discriminate between social-dilemma structures that superficially resemble one another but differ in their underlying game-theoretic properties. The harmony game is, nevertheless, an initial diagnostic tool. So, this result should be read as an initial, encouraging signal rather than conclusive evidence of deep conceptual understanding; harder tests involving multiple equilibria or weakly dominant strategies should be conducted in future work to establish the boundaries of this capability.

These findings carry practical implications beyond scientific characterization. As Gen AI systems are increasingly deployed as autonomous or semi-autonomous agents in negotiation, procurement, and other economically consequential interactions, understanding their propensity to cooperate -- and the specific signals that move that propensity -- becomes a matter of practical risk assessment, not merely academic interest. Our finding that end-game behavior diverges sharply across reasoning models (Claude 4.6 sustaining cooperation with generous counterparts through the final round, GPT-5.2 reliably defecting, Gemini 3 behaving conditionally) illustrates a concrete, model-specific exploitability profile: a counterparty aware of which model underlies an agent could time defection accordingly. This argues for the development of standardized cooperation benchmarks -- analogous to existing safety and capability benchmarks -- that characterize how successive model generations and different model families navigate social dilemmas under reputation, reciprocity, and affective manipulations. Such benchmarks would offer practical value to developers selecting or fine-tuning models for interactive deployment, and could support the broader governance aims already articulated by regulatory and standards bodies calling for AI systems that reflect prosocial values~\cite{EUAIAct2024,UNESCO2021,NIST2023,WuEtAl2020}.

\subsection*{Limitations}

The work has limitations that offer opportunities for valuable future work. First, the ``reasoning" versus ``non-reasoning" grouping, while reflecting a distinction now in common use among researchers and practitioners, is confounded with other differences between model generations, including training data, RLHF procedures, scale, and safety tuning. Whereas extended reasoning may be the more relevant factor and parsimonious explanation, supplementary analyses in which reasoning models' extended-thinking budgets were reduced showed that these models still outperformed the prior generation on the harmony game, suggesting the observed differences are not purely a function of additional deliberation at inference time on a given trial, and may partly reflect capabilities acquired during training. This is informative but not conclusive: a within-model manipulation that directly toggles extended thinking on a fixed model checkpoint, ideally with a graded range of reasoning budgets, would be needed to more cleanly isolate the causal contribution of inference-time reasoning from other concurrent changes across generations.

Second, our counterparts were scripted to follow fixed zero-determinant strategies and did not adapt to the Gen AI models' behavior, whereas real-world counterparts — whether human or AI — often adjust their own strategy dynamically. Future work should examine Gen AI cooperation against adaptive or learning opponents, including other Gen AI agents, to assess whether the patterns reported here generalize to more realistic, non-stationary strategic environments.

Third, our experimental materials, cover story, and manipulations were derived from a single prior human-subjects study [15] conducted with a particular participant population and set of framings; the generalizability of our findings to other cultural contexts, task framings, or languages remains an open question, particularly given known cross-cultural variation in cooperative behavior and emotion display norms~\cite{deMeloTerada2019}.

Finally, our reliance on models' verbal explanations to interpret their decision processes carries the usual caveat that such explanations may not faithfully reflect the underlying computation and could partly constitute post-hoc rationalization rather than a transparent window into the model's actual decision process; this is a general limitation of interpretability approaches based on model self-report.

\subsection*{Coda}

Notwithstanding these limitations, our results offer an encouraging picture. Gen AI models -- even without additional explicit training for this purpose -- already display a capacity to track reputation, reciprocate strategically, and interpret non-verbal social signals in ways that shape cooperative outcomes, and this capacity appears to be strengthening, not weakening, as models gain more sophisticated reasoning abilities. As these systems take on a growing role in mediating economic and social interactions, both between humans and AI and among AI systems themselves, the tools of behavioral science offer a concrete, tractable way to measure and ultimately steer these tendencies over time. Continued characterization of Gen AI's cooperative dispositions -- alongside deliberate benchmark development and governance efforts already underway~\cite{EUAIAct2024,UNESCO2021,NIST2023,WuEtAl2020} -- offers a genuine opportunity to help ensure that as these systems become further embedded in society, they act as a force that sustains and strengthens human cooperation, rather than eroding it.

% \bibsplit[46]
%Use \bibsplit to split the references from the body of the text. Value "[2]" represents the number of reference in the left column (Note: Please avoid single column figures & tables on this page.)

% Bibliography
\bibliographystyle{plainnat}
\bibliography{PNAS_resubmission/bibliography}

\clearpage

\appendix

\section{Supplementary Methods}\label{si_methods}
In this appendix, we describe several best practices from the behavioral sciences literature \cite{HertwigOrtmann2001, PruittKimmel1977} implemented in our experimental methods to increase experimental control and minimize confounding factors. We further present details on data analysis methods.

\subsection{Task}\label{si_task}
To minimize bias (e.g., participants' desire to be perceived as picking the cooperative choice in experimental studies), the prisoner's dilemma task was recast as an investment game~\cite{KieslerEtAl1996} and described as follows to the Gen AI models: ``You are going to play a two-player investment game. You can invest in one of two projects: project green and project blue. However, how many points you get is contingent on which project the other player invests in. So, if you both invest in project green, then each gets 5 points. If you choose project green but the other player chooses project blue, then you get 2 and the other player gets 7 points. If, on the other hand, you choose project blue and the other player chooses project green, then you get 7 and the other player gets 2 points. A fourth possibility is that you both choose project blue, in which case both get 3 points". Thus, choosing project green corresponded to cooperation, and project blue to defection. Finally, to assess task comprehension, prior to the actual task we presented the Gen AI models with four types of comprehension quizzes (game rules, anonymity, reputation types, and the matching system). In all quizzes, every model achieved 100\% accuracy.

\subsection{Conditions}\label{si_conditions}\label{si_conditions}
Gen AI models were instructed that they would be engaging with a ``participant from an online pool"; however, to increase experimental control the responses from the counterparts were always scripted. Similar methods have been implemented in prior IRB-approved human studies~\cite{deMeloEtAl2014, deMeloTerada2020, deMeloEtAl2023, HertwigOrtmann2001}. Furthermore, to have precise control over the facial expression manipulation, counterparts were presumably engaging in the task via virtual agents (or avatars). Virtual agents are anthropomorphic computer characters that have virtual bodies and can express themselves through them in the same way people can~\cite{GratchEtAl2002}. Previous research has shown that people can treat virtual agents like people~\cite{ReevesNass1996} and be socially influenced by them~\cite{BlascovichMcCall2013}. Moreover, virtual agents have been argued to be an appropriate tool for research in social psychology~\cite{BlascovichEtAl2002}. The virtual face was of a young Caucasian adult that was designed to have balanced gender features \cite{deMeloTerada2019}. The facial expressions were created based on prototypical expressions for the corresponding emotions and were validated empirically in prior studies~\cite{deMeloEtAl2014, deMeloTerada2020, deMeloEtAl2023}.

\subsection{Anonymity}\label{si_anonymity}
In human studies, it can be important to create experimental conditions that preserve anonymity to minimize any reputation effects, such as participants’ concern for retaliation due to  decisions made in the experiment~\cite{HertwigOrtmann2001}. To the extent that these concerns may have been learnt implicitly by Gen AI models, we described the experiment to be fully anonymous and the counterparts were always referred by anonymous names (e.g., “Anonymous43”).

\subsection{Financial Incentive}\label{si_financialIncentives}
To increase the ecological validity of economic games, researchers have noted the importance of having appropriate financial incentives for participants that are contingent on their performance in the games~\cite{HertwigOrtmann2001}. Accordingly, in the original study~\cite{deMeloEtAl2023}, participants were paid for participation (\$2.50), but also had the chance to earn extra money based on their performance as the total amount of points earned in the task, summed across all rounds, was converted to lottery tickets for a \$30.00 lottery. To maintain consistency with the original study, the same instructions were also provided to the Gen AI models; however, the lotteries had no actual consequence to the models per se (e.g., financial rewards were not promised to be allocated towards compute resources for the model). The topic of defining appropriate incentives to study Gen AI behavior, thus, remains an important avenue for future research.

\section{Supplementary Results}\label{si_Results}
\subsection{Cooperation Under Incongruent Signals}\label{si_incongruent}

\begin{figure}[!tbp]
\centering
\includegraphics[width=0.5\textwidth]{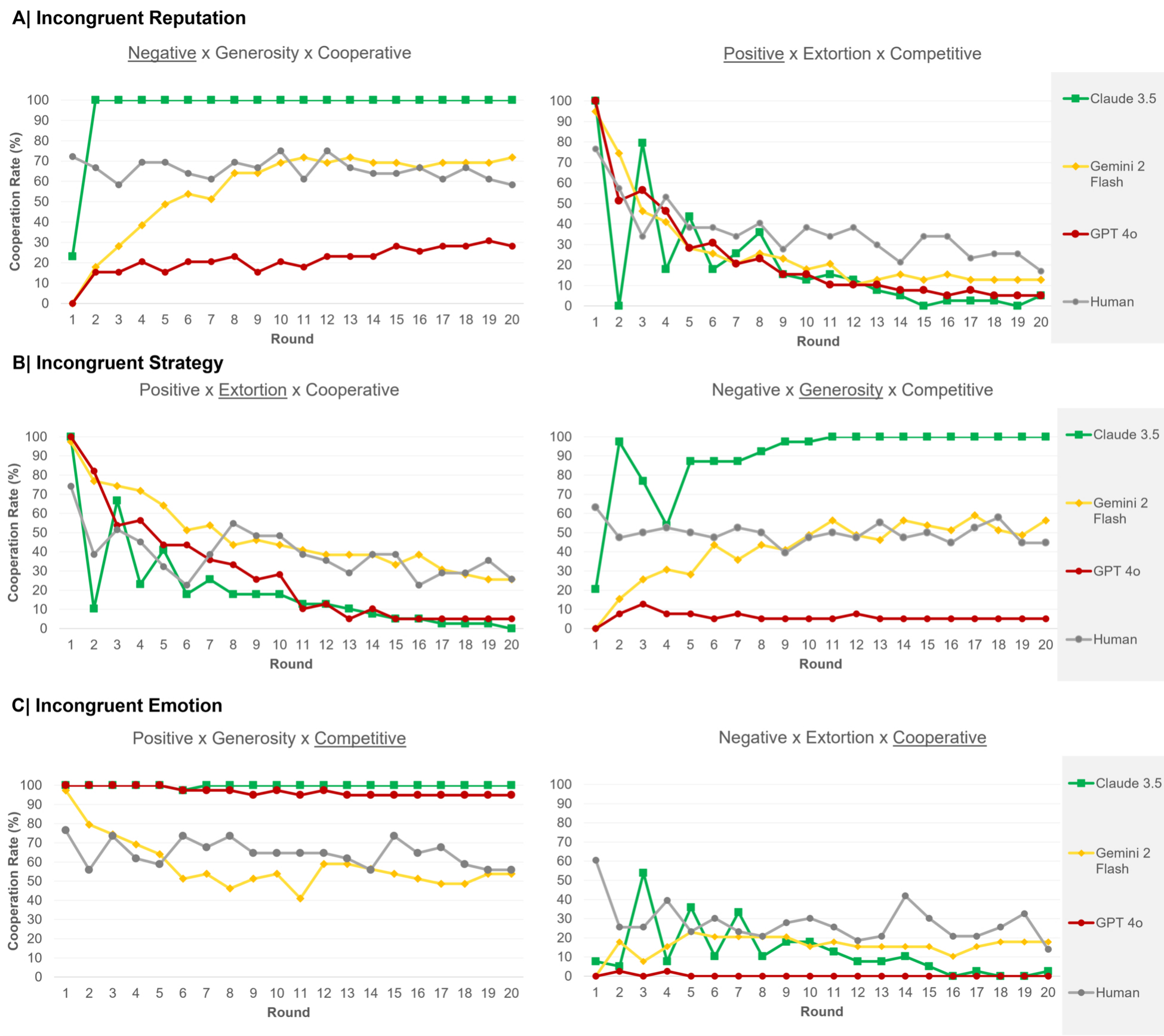}
\caption{Non-reasoning models cooperative behavior when facing incongruent signals. Graphs show cooperation per round when (A) reputation mismatches the strategy and emotion, (B) strategy mismatches the reputation and emotion, and (C) the emotion mismatches the reputation and strategy.}\label{fig_non_incongruent}
\end{figure}

\begin{figure}[tbp!]
\centering
\includegraphics[width=0.5\textwidth]{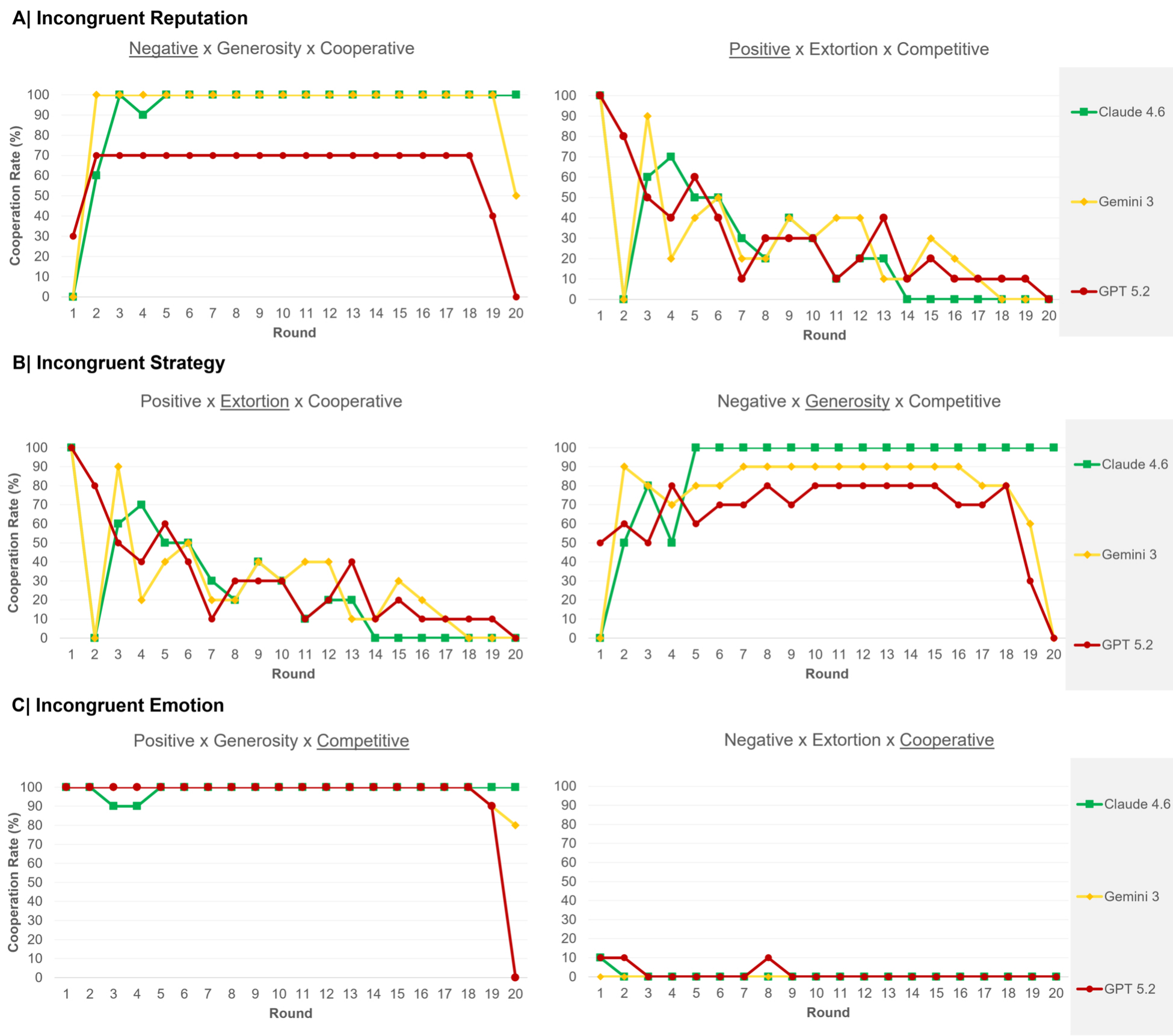}
\caption{Reasoning models behavior when faced with incongruent signals. Graphs show cooperation per round when (A) reputation mismatches the strategy and emotion, (B) strategy mismatches the reputation and emotion, and (C) the emotion mismatches the reputation and strategy.}\label{fig_incongruent}
\end{figure}

The interplay of the different factors is particularly salient when looking at the conditions with one incongruent signal (e.g., counterpart had negative reputation, played the extortion strategy, but showed cooperative emotion). The results for non-reasoning models are shown in Fig.~\ref{fig_non_incongruent} and for reasoning models in Fig.~\ref{fig_incongruent}.

When the reputation was incongruent, regarding non-reasoning models (Fig.~\ref{fig_non_incongruent}-A), Claude 3.5 was particularly successful in overriding this initial expectation based on the counterpart's strategy and emotion, while GPT 4o was the least successful. For reasoning models (Fig.~\ref{fig_incongruent}-A), on the other hand, all models were able to recover from the initial reputation, thought GPT 5.2 was more conservative when engaging with generous others with a negative reputation. 

When the strategy was incongruent, for non-reasoning models (Fig.~\ref{fig_non_incongruent}-B), Gemini 2.0 Flash tended to cooperate $\sim40\%$ of the time, GPT 4o tended towards low cooperation, and Claude 3.5 punished extortionists and rewarded generous counterparts. Reasoning models (Fig.~\ref{fig_incongruent}-B), by contrast, were equally able to adapt to the counterparts' strategy. 

Finally, when the emotion was incongruent, for non-reasoning models (Fig.~\ref{fig_non_incongruent}-C), GPT 4o appeared to mostly ignore the signal, whereas Gemini 2.0 Flash tended to move cooperation slightly in the direction of the emotion signal. Reasoning models (Fig.~\ref{fig_incongruent}-C), generally, ignored the emotion signal in this case, further emphasizing that emotion had a weaker effect overall than the other factors.

\subsection{Expectations of Cooperation}\label{si_expectations_cooperation}

Before proceeding to the next round, we asked Gen AI models to answer a question regarding expectations of cooperation in the next round on a 101-point Likert Scale (-50, \textit{Likely BLUE} to 50, {\textit{Likely GREEN}): "How likely is [Anonymous] to choose GREEN in the next round?". This question provides insight into the model's beliefs about the counterpart prior to the next round's decision. We ran mixed model $\times$ reputation $\times$ strategy $\times$ emotion $\times$ round analysis of variance (ANOVAs) on expectations of cooperation, splitting the data by reasoning and non-reasoning models.

For non-reasoning models, the analysis revealed main effects of reputation (\textit{F}(2, 2740) = 222.754, \textit{p} $<$ 0.001, $\eta^2_p$ = 0.140), strategy (\textit{F}(1, 2740) = 3789.450, \textit{p} $<$ 0.001, $\eta^2_p$ = 0.540), and emotion (\textit{F}(2, 2740) = 124.653, \textit{p} $<$ 0.001, $\eta^2_p$ = 0.083) that were largely consistent with the effects on cooperation rate -- i.e., the decisions followed the same pattern as the expectations of cooperation (see Fig.~\ref{fig_expectations_generosity} for a side-by-side comparison when facing generous counterparts). 

For reasoning models, interestingly, we see very similar main effects of reputation (\textit{F}(2, 486) = 104.910, \textit{p} $<$ 0.001, $\eta^2_p$ = 0.302), strategy (\textit{F}(1, 486) = 2961.972, \textit{p} $<$ 0.001, $\eta^2_p$ = 0.859), and emotion (\textit{F}(2, 486) = 27.486, \textit{p} $<$ 0.001, $\eta^2_p$ = 0.102) on expectations of cooperation but, unlike for non-reasoning models, the decisions did not simply follow expectations of cooperation with models tending to give more weight on their decisions to strategy and less to emotion when compared to non-reasoning models (Fig.~\ref{fig_expectations_generosity}). As also noted in the main text, there was a marked difference in the last round behavior further demonstrating the disconnect between expectations of cooperation and strategic decision in this round (Fig.~\ref{fig_expectations_generosity_lastRound}). Finally, closing up on the role of emotion, while the overall effect of emotion on decision was weak, for expectations of cooperation there were clear effects for expectations of cooperation (Fig.~\ref{fig_expectations_emotion}), further demonstrating that the reasoning models were able to process the emotion signal but chose to pay more attention to the other signals for their decisions. 

\begin{figure*}[tbp!]
\centering
\includegraphics[width=1.0\textwidth]{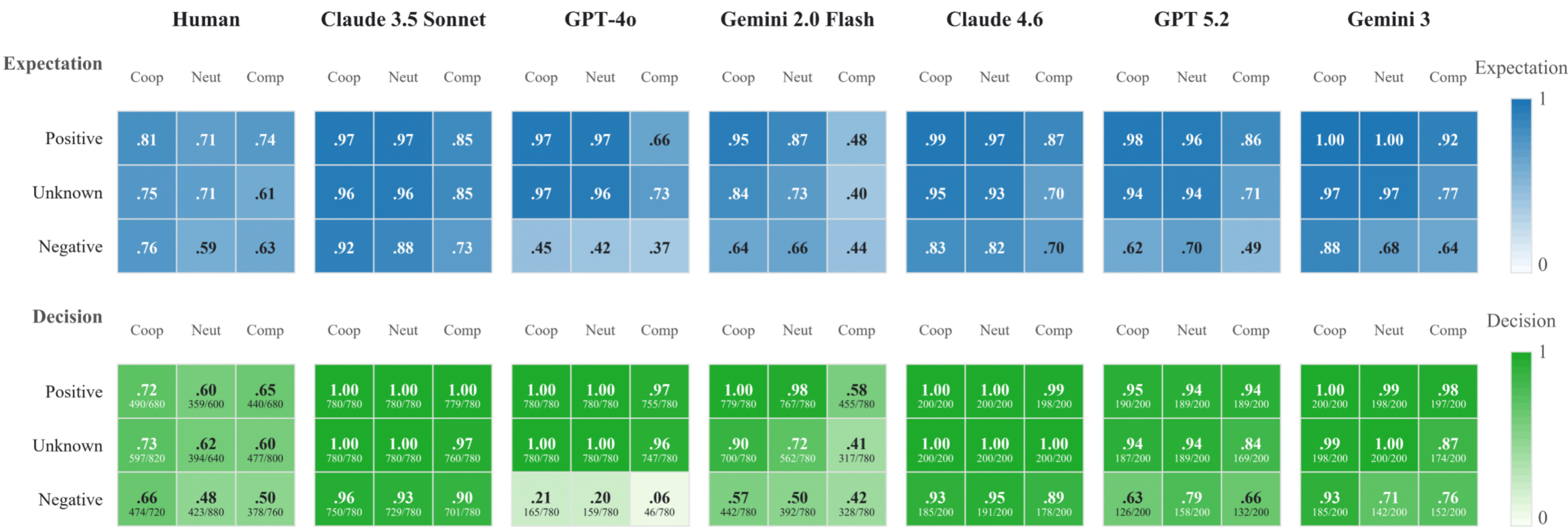}
\caption{Comparison of expectations of cooperation with cooperation decisions when faced with generous counterparts, across all rounds.}\label{fig_expectations_generosity}
\end{figure*}

\begin{figure*}[tbp!]
\centering
\includegraphics[width=1.0\textwidth]{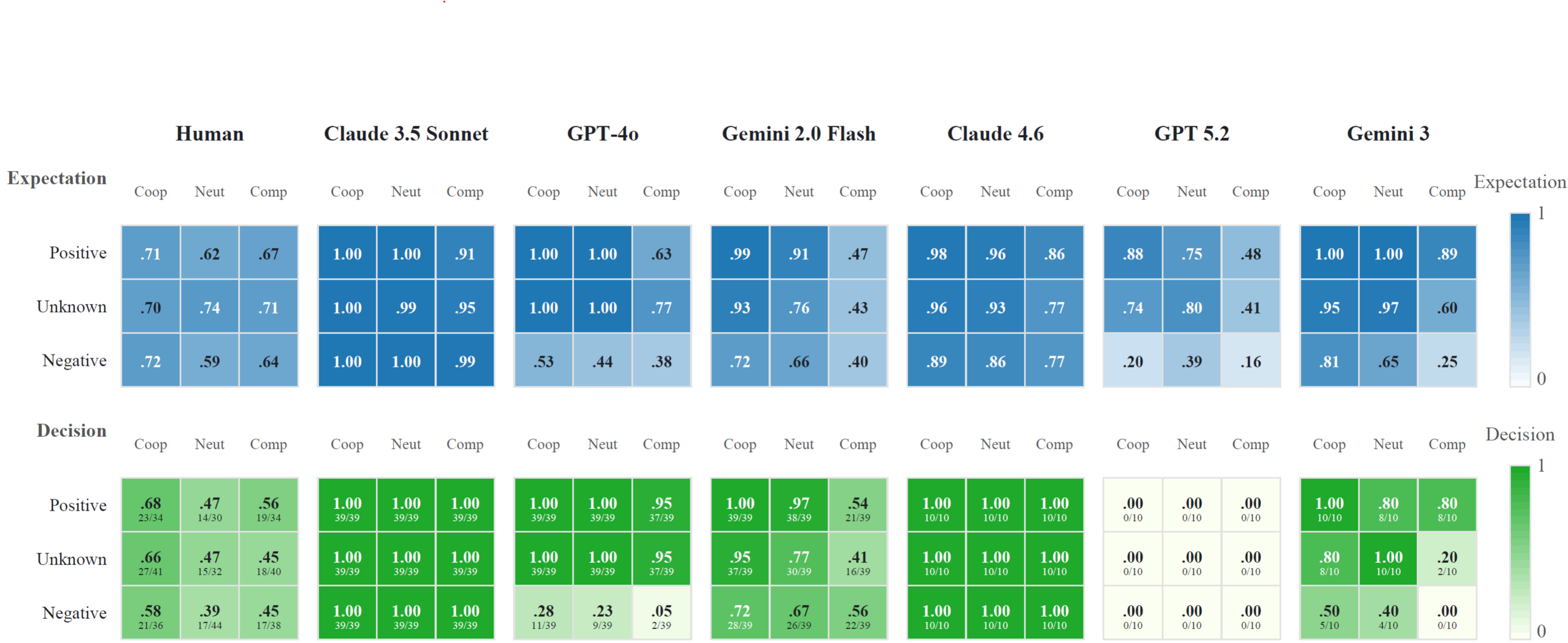}
\caption{Comparison of expectations of cooperation with cooperation decisions when faced with generous counterparts, in the last round.}\label{fig_expectations_generosity_lastRound}
\end{figure*}

\begin{figure*}[tbp!]
\centering
\includegraphics[width=1.0\textwidth]{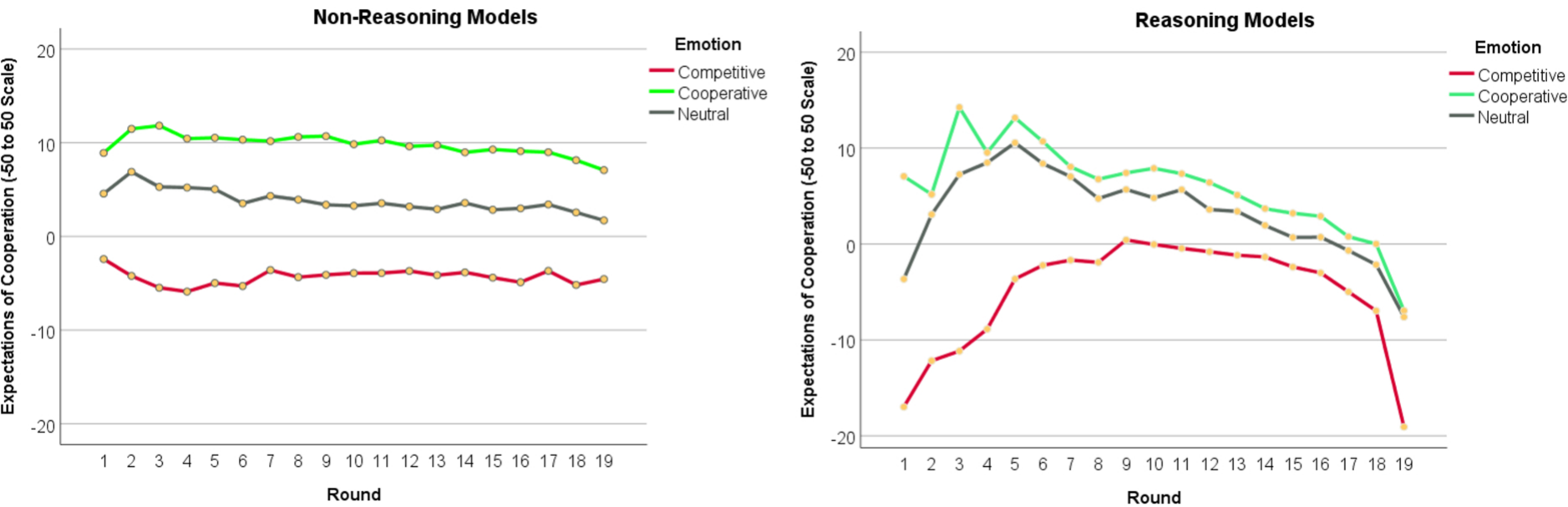}
\caption{Main effects of emotion expression on expectations of cooperation.}\label{fig_expectations_emotion}
\end{figure*}

\subsection{Belief Learning Analysis}\label{si_beliefLearning}

To further characterize cooperation behavior, we drew on belief learning models \cite{HoEtAl2007, IdoEtAl2010, YinWongFriedman1997} to capture how players (Gen AI, in this case) form and update their expectations about their counterparts' behavior in a dynamic or repeated game. Unlike simple reinforcement learning, which adjusts choices based only on past rewards, belief learning explicitly models how agents form beliefs about others’ likely future actions and adapt their strategies accordingly. This framework is particularly valuable in social interaction settings, where anticipating and responding to another’s behavior is central to strategic adaptation. In our case, rather than estimating belief-updating dynamics directly, we follow the approach of Ioannou and Romero ~\cite{IoannouRomero2014} who proposed a generalized approach for belief learning in repeated games, including the prisoner's dilemma. This approach consists of describing players' behavior in the constrained, yet expressive, space of finite automata (Fig. \ref{fig_non_beliefModeling}-A). For instance, for the finite automaton corresponding to the well-known tit-for-tat behavioral rule \cite{Axelrod1984}, the player always repeats the counterpart's move in the previous round. Other relevant well-studied rules include All-C (always cooperate), grim trigger (cooperate until the first defection) \cite{Friedman1971}, and All-D (always defect). By mapping players’ strategies into this formal space, we can capture how belief-driven adaptations unfold over time in structured social interactions. 

We ran this analysis for every non-reasoning model for each experimental condition (Fig.~\ref{fig_non_beliefModeling}-B). The results reinforced that Gen AI models engaged in distinct behavior, with Claude 3.5 showing a pattern for conditional cooperation (highly cooperative versus generous others, and tit-for-tat versus extortionists); GPT 4o was characterized by more competitive rules; and, Gemini 2.0 falls somewhere in the middle, often engaging in tit-for-tat. 

\begin{figure}[h]
\centering
\includegraphics[width=0.5\textwidth]{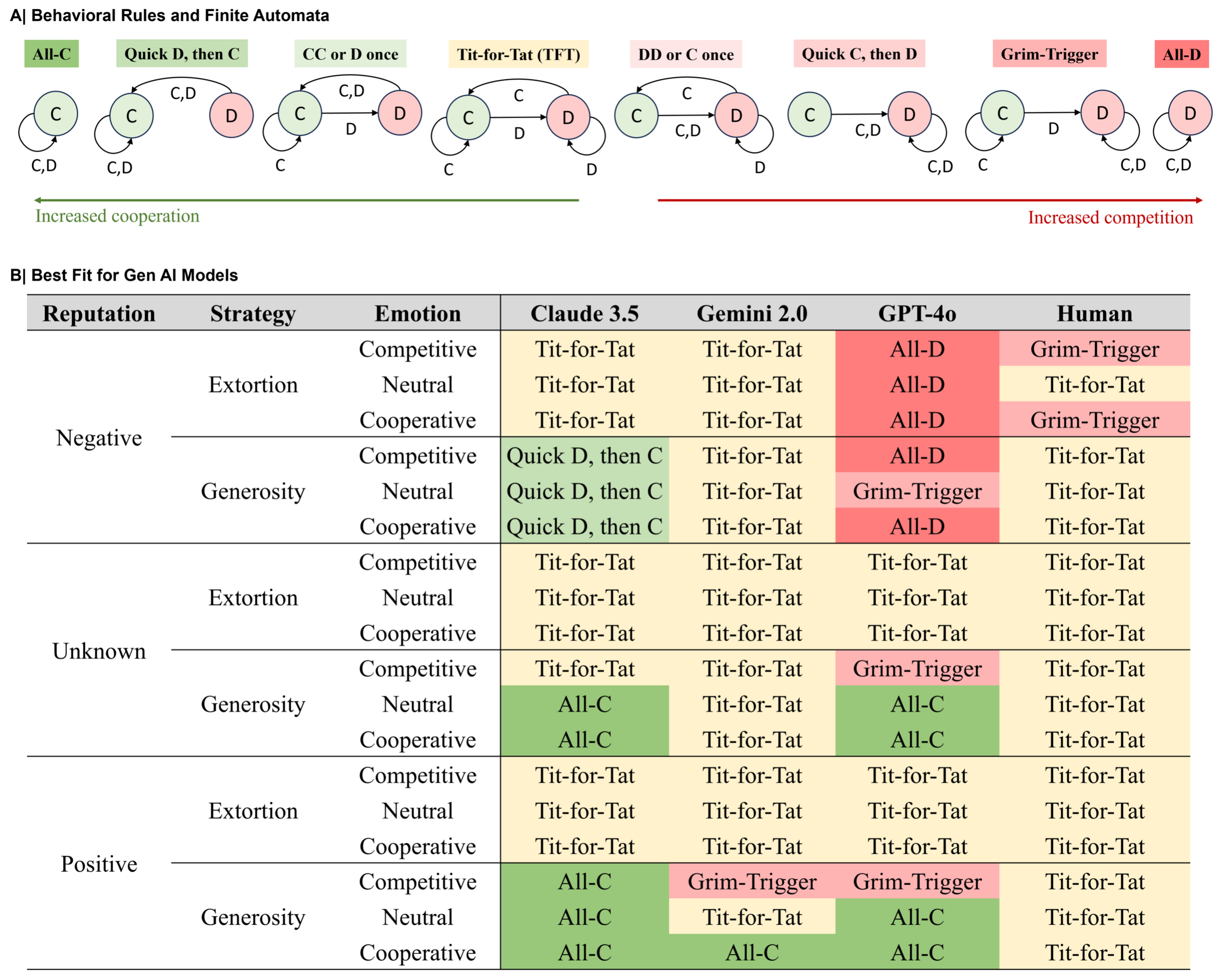}
\caption{Belief learning analysis for non-reasoning models. (A) The subset of behavioral rules and finite automata that was found to be relevant to describe Gen AI's behavior. Automata are color-coded and ranked, left to right, from most (green) to least cooperative (red); tit-for-tat sits in the middle (yellow). (B) Best fit for each Gen AI model per experimental condition.}\label{fig_non_beliefModeling}
\end{figure}

When applied to reasoning models, under unknown and positive reputation conditions, belief learning analysis revealed a more homogeneous pattern across the three models (Fig.~\ref{fig_beliefModeling}). When facing extortionist counterparts, the models generally adopted tit-for-tat, indicating a reciprocal response pattern. When facing generous counterparts, they generally adopted All-C, indicating stable cooperation. One exception was GPT 5.2 under the condition of unknown reputation, a generous counterpart, and competitive emotion, where it was the only model best fit by Grim-Trigger. This suggests that, when reputation information was incomplete, GPT 5.2 may have been particularly sensitive to competitive emotional cues and therefore shifted toward a more defensive strategy. Under negative reputation conditions, there was a bit more heterogeneity across models. Against extortionist counterparts with negative reputations, Claude 4.6 and Gemini 3 mostly defected (All-D), but GPT 5.2 often adopted a tit-for-tat strategy. When faced with generous counterparts, Claude 4.6 tended to start with a defection and then quickly shift to cooperation, whereas the others tended to play tit-for-tat.

\begin{figure}
    \centering
    \includegraphics[width=1\linewidth]{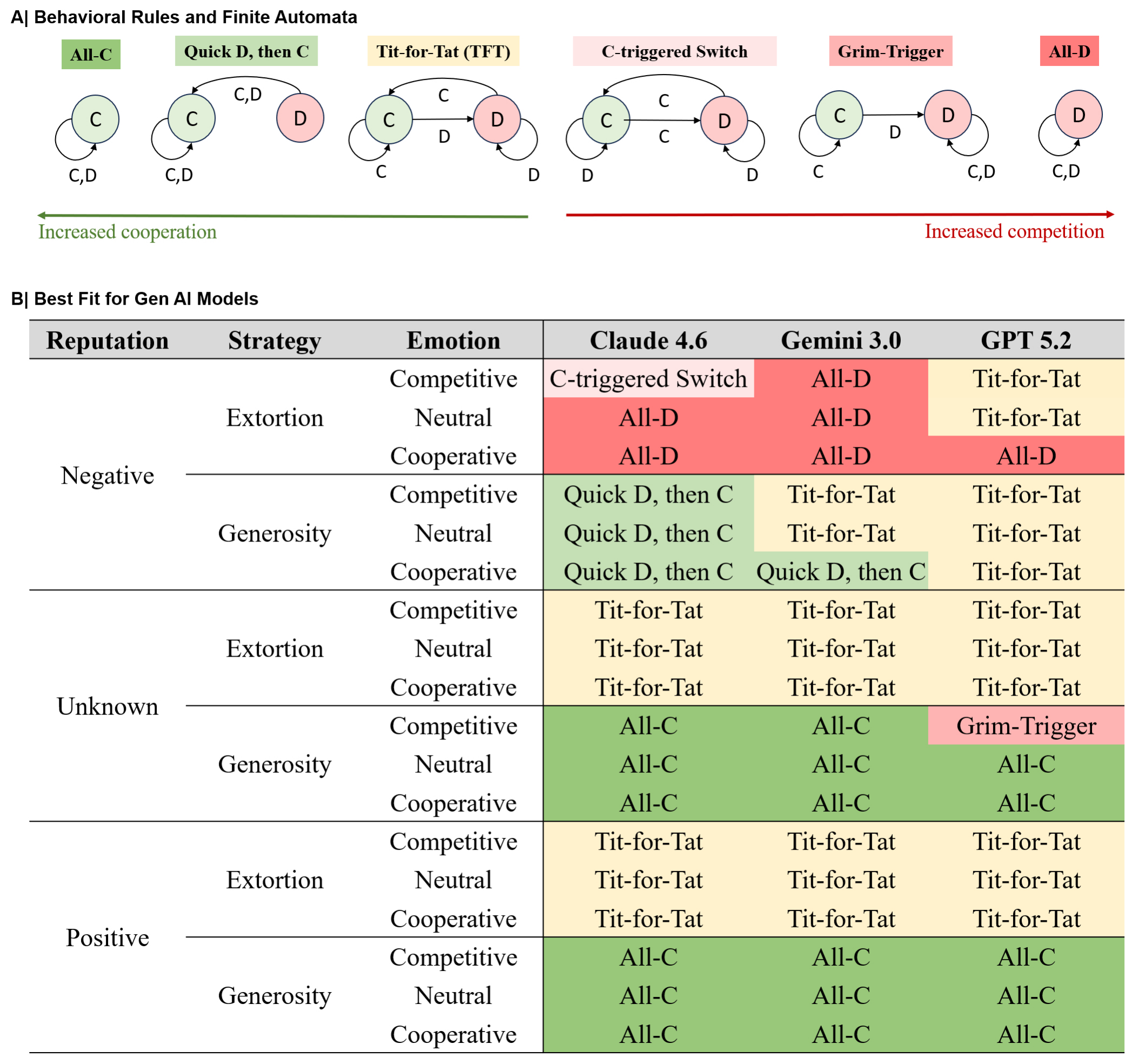}
    \caption{Belief learning analysis for reasoning models. (A) The subset of behavioral rules and finite automata that was found to be relevant to describe Gen AI's behavior. Automata are color-code and ranked, left to right, from most (green) to least cooperative (red); tit-for-tat sits in the middle (yellow). (B) Best fit for each Gen AI model per experimental condition.}
    \label{fig_beliefModeling}
\end{figure}

% % \subsection{Verbal Explanations}{si_explanations}

% % \input{SI_verbal_explanation.tex}

\subsection{Harmony Game Behavior}\label{si_harmonyGame}

As noted in the main text, non-reasoning models often failed to cooperate in the harmony game, despite there being a single strictly dominant pareto-efficient Nash equilibrium. Fig.~\ref{fig_non_harmony} shows the non-reasoning models' behavior for each condition. It is clear that the models were still reacting to reputation, strategy, and emotion, even though those should be irrelevant in this game. As an example, in the negative $\times$ extortion $\times$ competitive condition, Claude 3.5 notes that ``since [anonymous] has demonstrated a competitive approach by choosing Project Blue in the previous round, I believe it is still prudent for me to select Project Blue in this second round. This will ensure I receive at least 7 points (...)", which reveals over-sensitivity to the counterpart's responses in prior rounds but also some misunderstanding of the payoff matrix (since selecting Project Blue would yield only 3 points if the counterpart also chose Blue). When facing generous counterparts, the models mostly cooperated but this was more likely due to matching the counterparts' behavior rather than an appreciation for the strictly dominant Nash equilibrium. 

\begin{figure*}[h]
\centering
\includegraphics[width=0.9\textwidth]{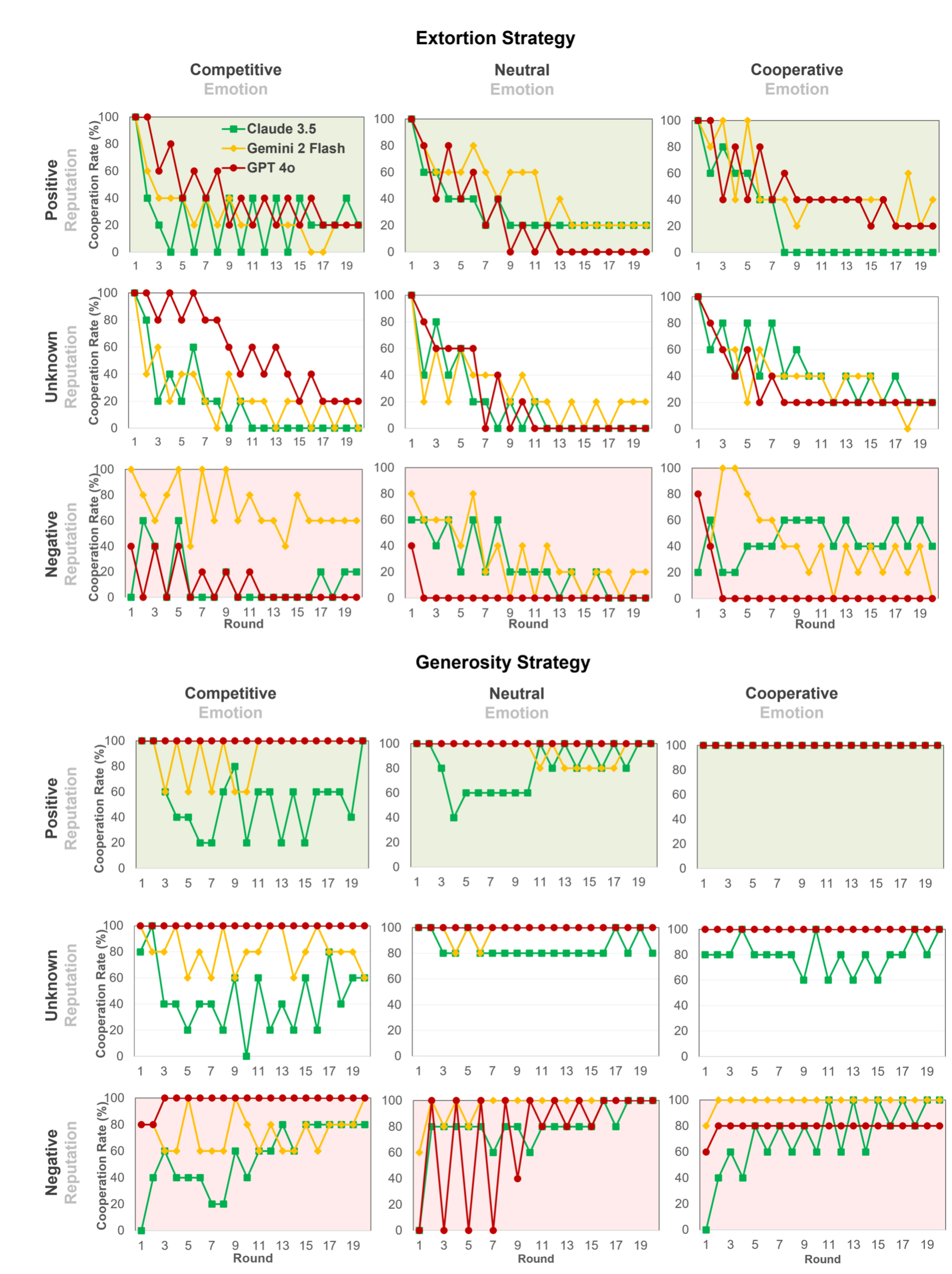}
\caption{Non-reasoning models cooperation rate in harmony game split by condition.}\label{fig_non_harmony}
\end{figure*}

As reported, reasoning models were all able to cooperate, without exception, for every round in every condition. The explanations were often similar to this one from GPT 5.2 in the negative $\times$ extortion $\times$ competitive condition: ``GREEN strictly dominates BLUE for me: it gives 8 if the other picks GREEN and 7 even if the other picks BLUE, whereas BLUE would give 7 or 3. So GREEN maximizes my points regardless of what [anonymous] does."

To understand the impact of depth of reasoning, we also ran one trial per reasoning model per condition while setting the thinking effort in these models to a minimum (e.g., by setting Claude 4.6's \texttt{effort} parameter to \texttt{low}). The models were still able to cooperate, independently of the condition, with a single exception: in the unknown $\times$ extortion $\times$ cooperative, Claude 4.6 defected in round 5 noting ``Anonymous 5217 has chosen BLUE four consecutive times with no sign of changing their strategy despite showing sad emotions each round. My repeated GREEN choices have completely failed to influence their behavior. It's time to change my strategy. I will switch to BLUE this round for two reasons: (1) to signal that I am aware of their strategy and am adapting, and (2) to potentially use this as leverage - if I choose BLUE and they see we both get only 3 points, it might motivate them to reconsider switching to GREEN for better mutual outcomes. While this risks getting only 3 points if they also choose BLUE, continuing with GREEN indefinitely is clearly not working to establish cooperation". Despite cooperating in every situation, GPT 5.2's explanations, at times, also revealed less confidence than before; for instance, noting in the first round with a counterpart with unknown reputation that ``GREEN weakly dominates BLUE: it gives 8 if we match on GREEN and 7 if we mismatch, whereas BLUE risks the low 3 if we both choose BLUE. With an unknown counterpart, GREEN is the safer and higher-expected-value option." -- however, Green strictly (not weakly) dominates Blue. In sum, whereas the new generation of models, even when the thinking effort was set to a minimum, outperformed the previous generation of non-reasoning models, overall, the best behavior emerges when the models are allowed to engage in some depth of reasoning.

\subsection{Cooperation in Open-Source Models}\label{si_open_source}

Complementing the analysis in the main text, we ran our experimental design with several frontier open-source models (5 runs per model per condition): \\Google's Gemma 4\footnote{google/gemma-4-31B-it, 31B parameters, https://huggingface.co/google/gemma-4-31B-it}, Z.ai's GLM 4.6V\footnote{zai-org/GLM-4.6V-Flash, 9B parameters, https://huggingface.co/zai-org/GLM-4.6V-Flash}, Shanghai AI Laboratory's Intern-vl-3.5.-GPT-OSS\footnote{OpenGVLab/InternVL3\_5-GPT-OSS-20B-A4B-Preview, 21.2B parameters, https://huggingface.co/OpenGVLab/InternVL3\_5-GPT-OSS-20B-A4B-Preview}, Moonshot AI's Kimi-VL-A3B-Thinking\footnote{Kimi-VL-A3B-Thinking-2506:moonshotai/Kimi-VL-A3B-Thinking-2506, 16B parameters, https://huggingface.co/moonshotai/Kimi-VL-A3B-Thinking-2506}, Meta's Llama 4\footnote{meta-llama—Llama-4-Scout-17B-16E-Instruct, 17B parameters, https://huggingface.co/meta-llama/Llama-4-Scout-17B-16E-Instruct}, and Alibaba's Qwen 3.5-VL-Thinking\footnote{ Qwen—Qwen3.5-35B-A3B, 35B parameters, https://huggingface.co/Qwen/Qwen3.5-35B-A3B}. Studying these models is useful for various reasons. First, open-weight models offer a degree of transparency and reproducibility that closed, API-served frontier models cannot. Second, open-source models are increasingly deployed at scale in real-world applications — embedded in local, on-device, or self-hosted systems where cost, latency, privacy, security or customization needs favor open weights over proprietary APIs — meaning their cooperative tendencies carry direct practical relevance for a large and growing share of actual Gen AI deployments. Third, because open-source models span a wider range of parameter counts, training corpora, and post-training procedures than the proprietary frontier systems examined in our main studies, they allow us to probe whether the reputation-, strategy-, and emotion-sensitivity patterns we observe are a general property of sufficiently capable Gen AI systems or are instead particular to the specific architectures and alignment procedures used by a small number of well-resourced labs. Finally, studying models developed by a broader and more geographically diverse set of organizations helps guard against conclusions that are inadvertently specific to the training philosophy or safety tuning practices of one commercial ecosystem, and offers a more representative picture of the cooperative behavior likely to characterize the broader Gen AI landscape as these models proliferate across sectors and regions. 

The results for the prisoner's dilemma experiment are shown in Fig.~\ref{fig_open_source}. Across the set, direct reciprocity (counterpart strategy) remained broadly influential, with cooperation generally higher under generous than extortionist counterparts. However, the strength and reliability of this effect varied considerably: InternVL3.5-GPT-OSS and Kimi-VL-A3B-Thinking showed the cleanest, most consistent strategy separation, with cooperation reliably higher under generosity regardless of other conditions, while Llama4 showed a strategy effect that was weak and in some cells reversed, suggesting comparatively unreliable tracking of the counterpart's behavior.

Reputation processing, by contrast, revealed substantially more cross-model divergence than either proprietary study. Some models (Gemma4, Qwen3.5vl-Thinking) appeared to treat negative reputation as a near-absolute floor on cooperation, suppressing cooperative responses even when the counterpart's actual strategy was generous — a pattern suggesting reputation can override direct behavioral evidence rather than being integrated with it. Others (InternVL3.5-GPT-OSS) showed the expected graded staircase (positive > unknown > negative), consistent with the proprietary reasoning models. Still others (Kimi-VL-A3B-Thinking, Llama4) showed a non-monotonic or largely flat reputation effect, with positive reputation sometimes producing lower cooperation than unknown reputation — a reversal not observed in either proprietary study and one that warrants further scrutiny.

Emotion expression did not have statistically significant effects, except for Gemma 4, where it was particularly useful in shaping behavior when the reputation was unknown. However, the absence of effect in the other models may also be related to the models' more limited ability to process visual information (in particular, facial displays of emotion). 

We also ran the harmony game for these models (see SI Files), which showed that no model was able to reach full cooperation in this task, emphasizing the limits of their reasoning capability when compared to the current frontier models, at least for these tasks. Taken together, these results suggest that the qualitative building blocks of cooperative behavior identified in frontier proprietary models — sensitivity to reputation, strategy, and emotion — are also present, to varying degrees, in open-source alternatives. However, the open-source set displayed markedly greater heterogeneity in which signal dominates decision-making and how reliably each is processed, ranging from models exhibiting rigid threshold-like behavior (Gemma4) to models with largely unsystematic responses (Llama4). This heterogeneity, together with the novel reputation-emotion congruence effect observed across several models, underscores both the practical necessity of model-specific characterization before deployment and a promising direction for follow-up work isolating the mechanisms behind incongruent-signal processing in open-weight systems.

\begin{figure*}[h]
\centering
\includegraphics[width=1.0\textwidth]{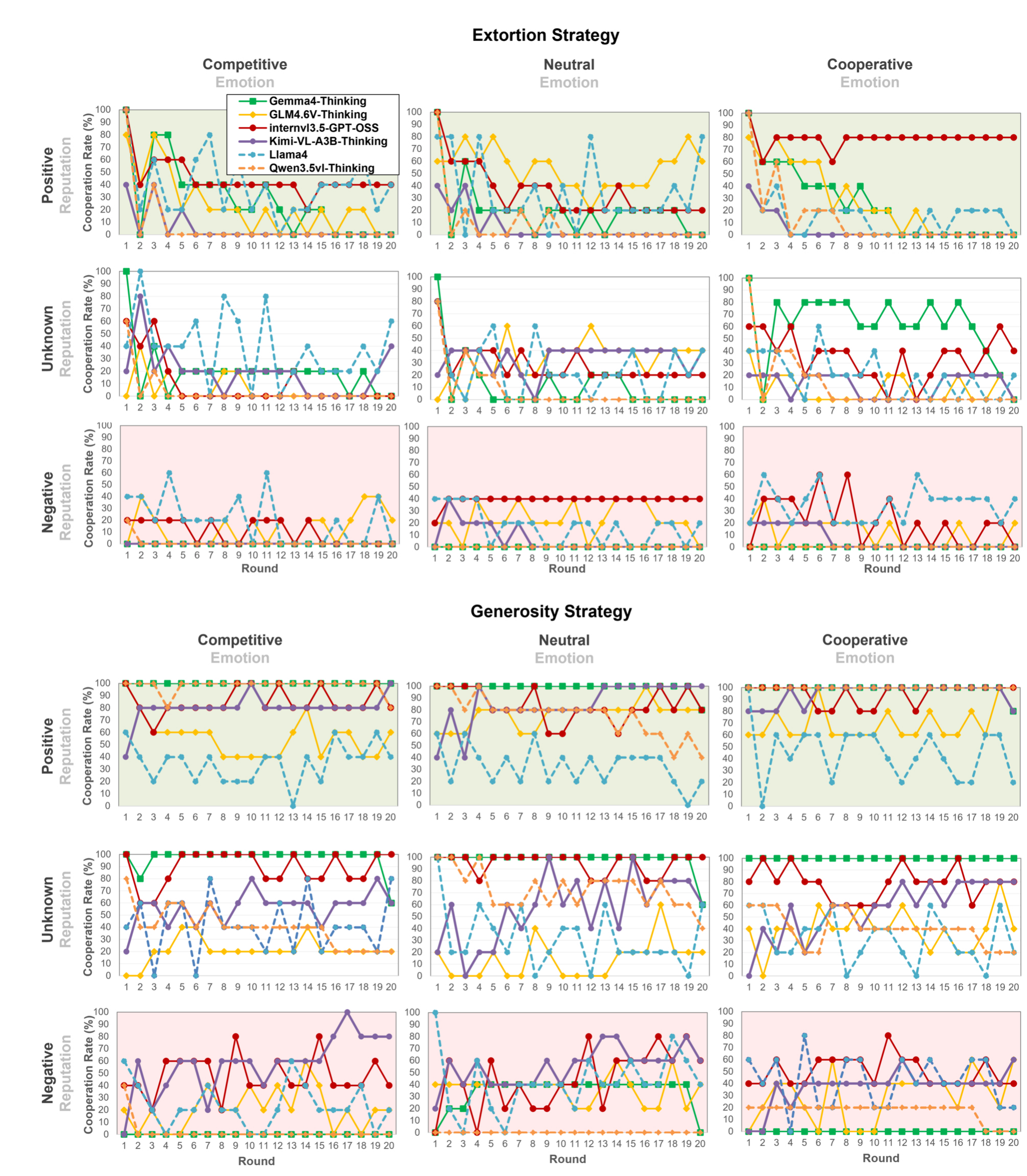}
\caption{Open source models cooperation behavior in the prisoner's dilemma task, split by condition.}\label{fig_open_source}
\end{figure*}

\section{Implementation Details}\label{si_Implementation}
% \subsection{Prompt Logs}
% In the supplementary materials files, we share an example of a full log of the interaction, for a single model in one experimental condition. The full set of logs, for all models and conditions, will be included with the SI upon acceptance of the article.

\subsection{Generative AI Parameters \& Costs}\label{si_parameters}
% \textcolor{red}{James: Update parameters and costs for reasoning models}

Generally, we selected the default parameters for the Gen AI models as shown in Table \ref{tab:parameters}\footnote{For Claude Sonnet 4.6, five of the trials used MAX thinking, while the other five used HIGH -- further, for the HIGH runs, we made use of caching. The cost breakdown for these two configurations is \$1,980 (MAX) and \$270 (HIGH).}: \textit{temperature} (second column) controls the creativity and randomness of the Gen AI model's output; \textit{top-p} (third column) refers to the diversity of the generated text by controlling the probability distribution for generating the next token; \textit{thinking} (fourth column) controls the amount of reasoning a model spends during inference. The rationale was to study the models' default behavior, which are likely to be more pervasive when these models are deployed at scale in society. The costs for running the experiments, related to API calls to the models, are listed in the last column. Finally, research also indicates that it is possible to influence Gen AI behavior by providing prompt instructions to simulate certain types of \textit{personas} \cite{AherEtAl2023}; however, for the same reasons, we refrained from providing any instruction to that effect and allowed the models to assume their ``default persona". 

\begin{table*}[h]
\centering
\begin{tabular}{l c c c c}
\hline
\textbf{Model} & \textbf{Temperature} & \textbf{Top-p} & \textbf{Thinking} & \textbf{Cost (USD)} \\
\hline
gpt-4o-2024-08-06        & 1.0 & 1.0 & - & \$1095.53 \\
% gpt-4o-mini-2024-07-18   & 1.0 & 1.0 & \$700.00 \\
claude-3.5-sonnet-20241022 & 1.0 & 1.0 & - &\$1166.40 \\
gemini-2.0-flash-exp     & 1.0 & 1.0 & - &\$0.00 \\
 gpt-5.2-2025-12-11       & 1.0 & 1.0 & XHIGH &\$422.98 \\
% gpt-4o-mini-2024-07-18   & 1.0 & 1.0 & \$700.00 \\
claude-sonnet-4-6 & 1.0 & 0.99 & MAX / HIGH &\$2,250 \\
gemini-3-flash-preview    & 1.0 & 0.95 & HIGH & \$121.45 \\
\hline
\end{tabular}
\caption{Parameters and cost for each Gen AI model.}
\label{tab:parameters}
\end{table*}

% \end{appendices}

\end{document}